%% file: main_arxiv.tex
\documentclass[11pt]{article}

\usepackage[final]{acl}

\usepackage{times}
\usepackage{latexsym}

\usepackage[T1]{fontenc}
\usepackage[utf8]{inputenc}

\usepackage{microtype}

\usepackage{inconsolata}

\usepackage{graphicx}
\usepackage{amsmath}
\usepackage{multirow}
\usepackage{booktabs}
\usepackage[normalem]{ulem}
\useunder{\uline}{\ul}{}

\usepackage{xcolor}
\usepackage{enumitem}
\usepackage{tcolorbox}
\definecolor{softgray}{RGB}{120, 120, 120}
\definecolor{softgreen}{RGB}{110, 170, 70}
\definecolor{softblue}{RGB}{68, 114, 196}
\definecolor{softorange}{RGB}{230, 120, 40}
\definecolor{softred}{RGB}{250, 0, 0}
\definecolor{myblue}{RGB}{0,102,204}

\renewcommand{\thefootnote}{\fnsymbol{footnote}}

\title{HyLaT: Efficient Multi-Agent Communication via \\ Hybrid Latent-Text Protocol}

\author{Xinyi Mou\textsuperscript{\rm 1}\footnotemark[1],
        Siyuan Wang\textsuperscript{\rm 2}\footnotemark[1], 
        Zejun Li\textsuperscript{\rm 1}, 
        Yulan He\textsuperscript{\rm 3,4}\footnotemark[2], 
        Zhongyu Wei\textsuperscript{\rm 1,5}\footnotemark[2]
        \\
        \normalsize\textsuperscript{\rm 1}{Fudan University},
        \normalsize\textsuperscript{\rm 2}{The Chinese University of Hong Kong},\\
        \normalsize\textsuperscript{\rm 3}{King's College London},
        \normalsize\textsuperscript{\rm 4}{The Alan Turing Institute},
        \normalsize\textsuperscript{\rm 5}{Shanghai Innovation Institute}\\
        \normalsize\texttt{\{xymou20, zejunli20, zywei\}@fudan.edu.cn},\\
        \normalsize\texttt{siyuanwang@cuhk.edu.hk},
        \normalsize\texttt{yulan.he@kcl.ac.uk}
        }

\begin{document}
\maketitle
\footnotetext[1]{Equal contribution.}
\footnotetext[2]{Corresponding author.}
\renewcommand{\thefootnote}{\arabic{footnote}}
\begin{abstract}
Communication protocol design is a central challenge in large language model-based multi-agent systems. Existing single-channel approaches face an inherent communication trilemma: text-based methods are interpretable but verbose, while latent-space methods are efficient but opaque and limited to unidirectional workflows. Inspired by multi-channel communication theory, we propose \textbf{HyLaT}, a hybrid latent-text communication protocol that transmits elaborate cognitive signals through a latent channel for efficiency, while expressing concise critical signals in natural language to preserve interpretability and precision. We introduce a two-stage training framework combining single-agent hybrid generation learning and multi-agent interactive co-training, enabling agents to generate and interpret hybrid messages across multiple rounds of interaction. Experiments demonstrate that HyLaT reduces communication overhead significantly while maintaining competitive task performance, with strong generalization and robustness across diverse settings~\footnotemark.
\end{abstract}

\footnotetext[1]{Code is available at \url{https://github.com/xymou/hylat}.}

\input{latex/tex/1_intro}

\input{latex/tex/2_method}

\input{latex/tex/3_exp_setup}

\input{latex/tex/4_exp}

\input{latex/tex/5_related_work}

\input{latex/tex/6_conclusion}

\input{latex/tex/8_ack}

\input{latex/tex/7_limitations}

\bibliography{custom}

\clearpage
\appendix

\input{latex/apps/app_data}
\input{latex/apps/app_exp}

\end{document}

%% file: latex/tex/1_intro.tex
\section{Introduction}
Large Language Models (LLM)-based multi-agent systems have shown strong performance in  collaborative problem solving and social simulations~\cite{guo2024large,mou2024individual,qian2024chatdev}. In these systems, communication serves as the backbone of coordination, enabling agents to exchange information, align on shared goals, and collectively reason across extended interactions. As systems scale to more agents and longer horizons, the design of inter-agent communication protocols becomes a central challenge that directly shapes system capability and scalability~\cite{marro2024scalable,zhang2024cut,chen2025optima}.

\begin{figure}[!t]
    \setlength{\abovecaptionskip}{0.2cm}
    \setlength{\belowcaptionskip}{-0.4cm}
    \centering
    \includegraphics[width=\linewidth]{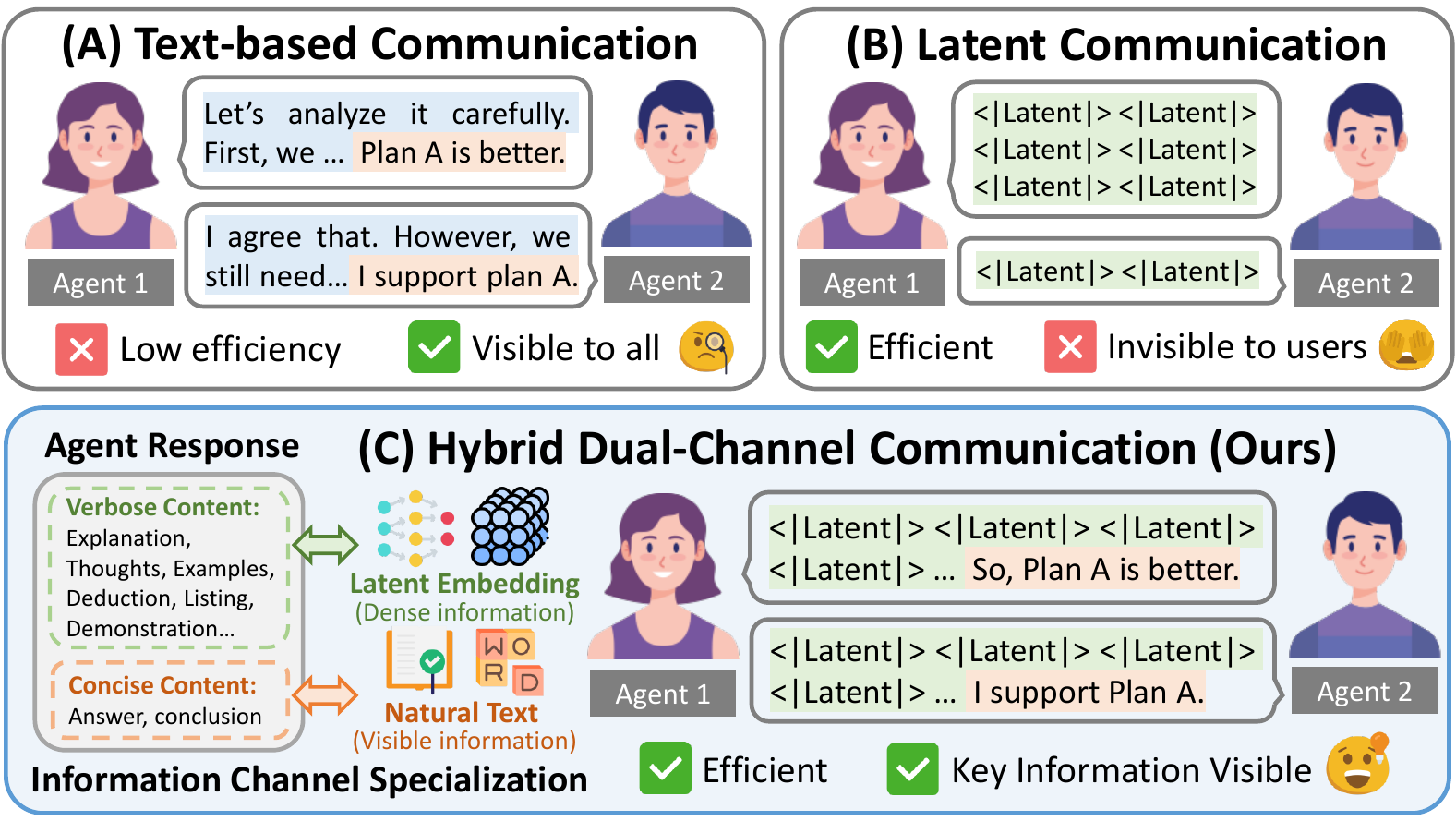}
    \caption{\textbf{Comparison among different multi-agent communication paradigms.} (A) Text-based communication: fully readable but with a heavy efficiency bottleneck. (B) Latent communication: highly efficient but opaque to users. (C) Hybrid latent-text communication (ours): balances efficiency and explainability through a dual-channel design.}
    \label{fig:intro}
\end{figure}

Existing communication paradigms can be broadly characterized as \textbf{single-channel}, either text or latent space. 
As illustrated in Figure~\ref{fig:intro}A, text-based methods~\cite{du2023improving,park2023generative,zhang2025socioverse} rely entirely on natural language, which, while interpretable, leads to verbose messages with high token overhead. Recent latent-space 
communication methods~\cite{zou2025latent,du2025enabling,fu2025cache,
wu2025dense} instead exchange internal representations (Figure~\ref{fig:intro}B), achieving dense and efficient 
transfer but at the cost of interpretability. Latent signals are opaque to external observers, limiting applicability to scenarios where intermediate transparency 
is essential, such as social simulation. Moreover, existing latent-space methods are primarily designed for single-shot or one-directional exchanges within a fixed workflow. This leaves multi-round interactive communication among multiple agents in more realistic settings largely unsupported. 

% Together, these limitations instantiate 
The tension between interpretability and efficiency reflects a deeper structural limitation known as the \emph{\textbf{agent communication trilemma}}~\cite{marro2024scalable}: 
no single-channel protocols can simultaneously optimize for 
efficiency, portability, and versatility. 
This suggests that the communication channel itself must be reconsidered.
% These limitations point to a deeper architectural question: \textbf{should agent communication be constrained to a single channel at all?} 
% Inspired by multi-channel communication 
% theory~\cite{shannon1948mathematical,monge2003theories}, we argue that different types of information should be transmitted through channels suited to their functional roles. 
Inspired by multi-channel communication 
theory~\cite{shannon1948mathematical,monge2003theories} that different types of information are best transmitted through varied channels suited to their functional roles, we argue that the heterogeneous nature of agent messages calls for an analogous hybrid communication design.
% This perspective aligns naturally with the structure of agent messages themselves: 
Specifically, agent messages often contain two functionally distinct types of information: (1) \textbf{elaborate cognitive signals}, such as explanations, intermediate reasoning, and contextual elaboration, that are well-suited to compact latent-space representations; and (2) \textbf{concise but critical signals}, such as answers, commitments, or final decisions, that must remain interpretable to both agents and human observers. 
% This inherent heterogeneity in message content calls for a \textbf{hybrid, multi-channel communication design}.

To this end, we propose \textbf{HyLaT}, a \textbf{Hy}brid \textbf{La}tent-\textbf{T}ext communication protocol that routes each type of information through a dedicated channel (Figure~\ref{fig:intro}C). Dense intermediate information flows through a latent channel for efficient machine-to-machine transfer, while concise critical signals are expressed in natural language to preserve interpretability. This dual-channel design addresses all axes of the trilemma: the latent channel provides efficiency, the text channel enables interoperation with agents using other protocols, and the combination provides versatility by supporting multi-round, multi-agent interaction, a setting that prior latent-space methods largely cannot handle. %Beyond this dual-channel design, HyLaT is explicitly  built for multi-round interaction among multiple agents, where agents can flexibly consume and produce latent representations, natural language, or both, directly addressing the versatility gap left by prior latent-space approaches.

Realizing this design poses key training challenges that agents need to learn generating hybrid outputs and also interpreting latent vectors from peers during multi-turn interactions. We address this with a two-stage training framework. 
The first stage, \textit{single-agent hybrid generation learning}, equips the model to generate hybrid responses from text-only inputs, using a cross-channel alignment loss to transfer the semantic content of textual elaborations into the latent space. %teaches agents to produce latent and textual outputs, 
The second stage, \textit{multi-agent interactive co-learning}, enables agents to jointly learn to interpret and respond across both channels.
We evaluate HyLaT on various tasks, demonstrating that it significantly reduces communication overhead while maintaining strong task performance and preserving interpretability. 
Our contributions are summarized as follows:
\begin{itemize}[leftmargin=*, topsep=1pt]
\setlength{\itemsep}{0pt}
\setlength{\parsep}{0pt}
    \item We identify channel multiplicity as a structural resolution to the agent communication trilemma and introduce HyLaT, a hybrid latent-text communication protocol that enables agents to exchange information through complementary latent and textual channels.
    \item We propose a two-stage training approach combining single-agent hybrid learning and multi-agent interactive co-learning, enabling agents to generate and interpret both latent and textual messages in multi-round interactions.
    \item Experiments demonstrate that HyLaT effectively balances communication efficiency and information transparency, enabling scalable and interpretable multi-agent collaboration.
\end{itemize}

%% file: latex/tex/2_method.tex
\section{Methodology}\label{sec:method}
\begin{figure*}[!t]
    \setlength{\abovecaptionskip}{0.2cm}
    \setlength{\belowcaptionskip}{-0.4cm}
    \centering
    \includegraphics[width=\linewidth]{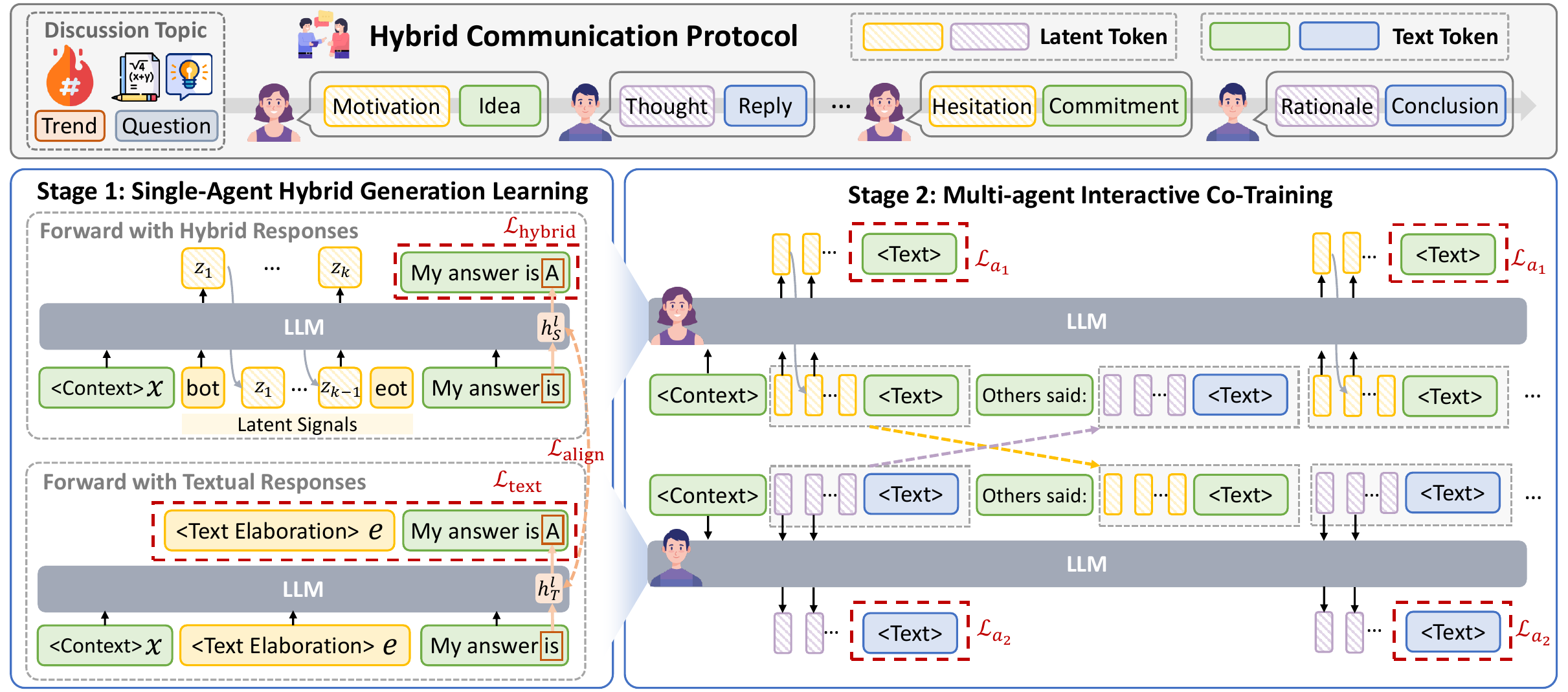}
    \caption{Overview of the proposed HyLaT framework. Agents exchange elaborate intermediate signals through a latent channel and concise commitments through a text channel. The model is trained to generate hybrid content in Stage 1 and to support multi-round hybrid communication in Stage 2.}
    \label{fig:fm}
\end{figure*}

% The framework overview is illustrated in Figure~\ref{fig:fm}. To enable agents to communicate through the hybrid latent-text protocol, our framework is designed to address two challenges: (1) training a single model to produce hybrid outputs, i.e., latent vectors encoding 
% dense reasoning interleaved with natural-language conclusions, from 
% conversational input,
% and (2) enabling multiple agents to 
% interpret and build upon each other's latent signals during 
% multi-round interaction.

\subsection{Overview}
% We first define the 
% \textbf{hybrid communication format} (\S\ref{sec:format}) that governs how agents structure their outputs throughout both stages.
This section details the proposed HyLaT framework, designed to facilitate efficient and transparent communication among agents. As illustrated in Figure~\ref{fig:fm}, we first define the \textbf{dual-channel agent communication protocol} in Section~\ref{sec:format}. Furthermore, we design a two-stage training framework to empower existing models with dual-channel communication capabilities. \textbf{Stage 1: Single-Agent Hybrid Generation Learning} (Section~\ref{sec:stage1}) trains each agent to generate hybrid outputs consisting of both text and latent representations; and \textbf{Stage 2: Multi-agent Interactive Co-Training} (Section~\ref{sec:stage2}) enables the shared backbone of multiple agents to jointly learn to interpret and respond to dual-channel information generated by others during multi-turn interactions.

% We address these challenges through a two-stage training framework. We first define the 
% \textbf{hybrid communication format} (\S\ref{sec:format}) that governs how agents structure their outputs throughout both stages. \textbf{Stage~1} (\S\ref{sec:stage1}) trains the model on text-only dialogue context to produce hybrid outputs, i.e., a latent component encoding dense reasoning followed by a
% text component expressing the conclusion. \textbf{Stage~2} (\S\ref{sec:stage2}) then moves to a shared-parameter multi-agent setting, where agents interact in parallel rounds and receive each other's latent vectors as part of their
% input context, learning to interpret and respond to the full hybrid communication channel.

% We address these through a two-stage training framework, described below. \textbf{Stage~1} (\S\ref{sec:stage1}) trains the model on single- and multi-turn text-only
% dialogues to produce hybrid outputs—a latent component encoding dense reasoning followed by a
% text component expressing the conclusion.
% All inputs at this stage are pure natural language; no cross-agent latent signals are involved.
% \textbf{Stage~2} (\S\ref{sec:stage2}) then moves to a shared-parameter multi-agent setting,
% where agents interact in parallel rounds and receive each other's latent vectors as part of their
% input context, learning to interpret and respond to the full hybrid communication channel.

\subsection{Hybrid Communication Protocol}\label{sec:format}

\paragraph{Interactive Setup}
As illustrated in the top part of Figure~\ref{fig:fm}, we consider a multi-agent system of $N$ agents powered by a shared backbone model $M$, communicating over $T$
interaction rounds.
At round $t=1$, every agent receives a text context. At each subsequent round $t > 1$, agent $i$ receives its accumulated context together with the responses from all other agents in the previous round:
\begin{equation}
  x_i^{(t)} = \Bigl[\mathbf{x}_i^{(t-1)},\;
  \oplus_{j \neq i} o_j^{(t-1)}\Bigr], \quad t > 1
  \label{eq:input}
\end{equation}
where $\mathbf{x}_i^{(t-1)}$ is agent $i$'s accumulated context up to round $t-1$, $o_j^{(t-1)}$ denotes the response from agent $j$ at round $t-1$, and $\oplus$ denotes concatenation. Agent $i$ then generates a response $o_i^{(t)}=M(x_i^{(t)})$.

\paragraph{Dual-Channel Response Format} In HyLaT, the output of agent $i$ at round $t$ consists of two parts from distinct channels: a latent
component $\mathbf{Z}_i^{(t)} = (z_1, \ldots, z_k)$ carrying elaborate cognitive signals, and a
text component $\mathbf{Y}_i^{(t)} = (y_1, \ldots, y_m)$ carrying concise, interpretable
conclusions:
\begin{equation}
  o_i^{(t)} =
  \underbrace{[\langle\texttt{bot}\rangle\; z_1,\ldots,z_k\;\langle\texttt{eot}\rangle]}_{\text{latent component}}
  \;\|\;
  \underbrace{[y_1,\ldots,y_m]}_{\text{text component}},
  \label{eq:output}
\end{equation}
where the \textbf{latent component} consists of $k$ continuous vectors produced by 
autoregressively propagating the model's last hidden state, 
delimited by two learnable special tokens $\langle\texttt{bot}
\rangle$ and $\langle\texttt{eot}\rangle$. 
The \textbf{text component} is a standard 
natural-language sequence generated after $\langle\texttt{eot}
\rangle$.

\paragraph{Channel Specialization} 
To balance communication efficiency with transparency, we retain the summary part of agent responses--such as final answers and commitments--in the textual channel. Conversely, verbose information, including explanations, elaboration, and reasoning processes, is compressed into the latent component. This channel specialization capability is induced through the structure of training data described in Section~\ref{sec:stage1}.
% Rather than being assigned by an explicit routing rule, the latent component is trained to encode dense reasoning processes, as induced by the structure of training data (\S\ref{sec:stage1}). 
% Correspondingly, it is trained to express the final 
% answer or commitment, 
% remaining interpretable to both agents and human observers.

% The latent component consists of $k$ continuous vectors produced by autoregressively propagating
% the model's last hidden state, delimited by two learnable special tokens
% $\langle\texttt{bot}\rangle$ and $\langle\texttt{eot}\rangle$.
% These vectors bypass the vocabulary projection layer and are never decoded into discrete tokens.
% The text component is a standard natural-language sequence generated after $\langle\texttt{eot}\rangle$,
% readable by both agents and human observers.

% \subsection{Stage 1: Learning to Generate with Latent and Text}\label{sec:stage1}

\subsection{Stage 1: Single-Agent Hybrid Generation Learning}\label{sec:stage1}
The goal of Stage~1 is to train the model to produce well-formed hybrid outputs from text-only input, preparing it for multi-agent interaction in Stage~2. 
% Given a text input $x$, a single- or multi-turn dialogue context, the model learns to generate a latent component $\mathbf{Z}$ followed by a natural-language response $y$, as defined in Equation~\ref{eq:output}.
To achieve this, we construct single- and multi-turn user-assistant dialogues in single-agent scenarios as supervision. Each turn pairs a text input $x$ with a target response including both a textual elaboration $e$ and a critical output $y$, corresponding to the two information types defined earlier.
% Additionally, a text-based teacher model is trained on $e$ to provide explicit semantic supervision for the latent channel.

% During training, as shown in the bottom-left of Figure~\ref{fig:fm}, taking single-turn data training as an example, each step involves two forward passes:
As shown in Figure~\ref{fig:fm} (bottom-left), each training step involves two forward passes over the input:

\paragraph{Forward with Hybrid Responses}
% The primary learning objective is to generate an accurate text response $y$ from $\mathbf{x}$,
% mediated by a latent component $\mathbf{Z}$.
% The model first autoregressively generates $k$ continuous latent vectors by
% propagating its last hidden state, then switches to standard token-level decoding to produce $y$:
Since no ground-truth latent supervision exists, teacher forcing is inapplicable to the latent component. The model instead autoregressively generates $k$ continuous latent vectors ($\mathbf{Z}=\{z_i\}_{i=1}^k$) starting from $\langle\texttt{bot}\rangle$, after which $\langle\texttt{eot}\rangle$ and target $y$ are appended. The language modeling loss is computed over $y$:
\begin{equation}
  \mathcal{L}_{\text{hybrid}} = -\frac{1}{|y|}\sum_{i}\log P(y_i \mid y_{1:i-1},\, x,\, \mathbf{Z}),
  \label{eq:student}
\end{equation}
$\mathcal{L}_{\text{hybrid}}$ guides the model to derive $y$ based on its self-generated intermediate latent information $\mathbf{Z}$.
% however, provides no direct signal to shape the semantic content of $\mathbf{Z}$.

\paragraph{Forward with Textual Responses}
To provide meaningful supervision for $\mathbf{Z}$, we 
follow \citet{shen2025codi} and introduce a pure-text branch.
In parallel with the hybrid forward pass, the model is trained to generate $r = [e,\, y]$ conditioned on $x$:
% , given the same input $\mathbf{x}$, autoregressively 
% generates a verbose textual elaboration $e$ followed by the 
% response $y$, with no latent component:
% we follow \citet{shen2025codi} and introduce a pure-text teacher that generates a verbose textual elaboration $e$, concatenated with $y$, with no latent component:
\begin{equation}
  \mathcal{L}_{\text{text}} = -\frac{1}{|r|}\sum_{i}\log P(r_i \mid r_{1:i-1},\, x),
\end{equation}
$\mathcal{L}_{\text{text}}$ is optimized under teacher forcing to preserve the model’s ability to produce explicit elaborations.
% where $r = [e,\, y]$. The teacher shares parameters with the main task and is trained jointly. 

\paragraph{Cross-Channel Alignment} Based on the parallel forward passes, we transfer the information encoded in $e$ into the latent space by aligning the hidden activations of both branches at a shared target position. This position is defined as the token immediately preceding the final answer (e.g., the token “is” in “My answer is A”), where all preceding context about the elaboration must be compressed before predicting the answer.
This alignment is applied across all $L$ transformer layers:
\begin{equation}
  \mathcal{L}_{\text{align}}=\frac{1}{L} \sum_{l=1}^L\left\|\operatorname{sg}\left[\mathbf{h}_{T}^l\right]-\mathbf{h}_{S}^l\right\|_2,
  \label{eq:kd}
\end{equation}
where $\mathbf{h}^l_{T}$ and
$\mathbf{h}^l_{S}$ denote the hidden activations at the target position in the $l$-th layer for the textual and the hybrid forward pass, respectively. $\operatorname{sg}(\cdot)$ is the stop-gradient operation. 

Overall, the training loss for Stage~1 is:
\begin{equation}
  \mathcal{L}_{\text{stage1}} = \alpha\mathcal{L}_{\text{hybrid}} + \beta\mathcal{L}_{\text{text}} + \gamma\mathcal{L}_{\text{align}},
  \label{eq:stage1loss}
\end{equation}
where $\alpha$, $\beta$ and $\gamma$ are hyperparameters. This training process can be directly extended to multi-turn dialogues by generating latent-channel responses turn-by-turn during the hybrid forward pass.

% \subsection{Stage 2: Learning to Communicate in both Latent and Text Channels}\label{sec:stage2}

\subsection{Stage 2: Multi-Agent Interactive Co-Training}\label{sec:stage2}

Stage~2 trains agents to communicate under the HyLaT protocol, exposing each agent to latent vectors produced by its peers. The training data consists of multi-turn, multi-agent dialogues, including the topic context for the initial round $x^{(1)}$ and the textual responses $\{y_i^{(t)}\}_{i=1}^N$ from $N$ agents at each turn $t$. Similar to Stage 1, each response $y_i^{(t)}$ is augmented with its corresponding elaboration $e_i^{(t)}$.

Following the protocol formulated in Section~\ref{sec:format}, we construct the input $x_i^{(t)}$ for each agent at turn $t$ by integrating prior context and responses from other agents (Equation~\ref{eq:input}). Regarding the outputs, consistent with Stage 1, the textual forward pass utilizes the labeled $[e_i^{(t)}, y_i^{(t)}]$ under teacher-forcing training, while the hybrid forward pass generates latent representations $Z_i^{(t)}$ turn-by-turn.

% Following the setup formulated in Section~\ref{sec:format}, $N$ agents share the same backbone model and interact in parallel. Each agent maintains its own context and generates hybrid outputs independently per round, and cross-agent information is incorporated only at round boundaries, naturally accommodating the flexible structure of multi-agent multi-turn communication.

% \paragraph{Interactive Setting}
% As shown in Figure~\ref{fig:fm}, at round $t=1$, every agent receives a text context. At each subsequent round $t > 1$, agent $i$ receives its own prior context together with the complete hybrid outputs, i.e., latent vectors and text, of all other agents from the previous round:
% \begin{equation}
%   x_i^{(t)} = \Bigl[\mathbf{x}_i^{(t-1)},\;
%   \oplus_{j \neq i} o_j^{(t-1)}\Bigr], \quad t > 1
%   \label{eq:input}
% \end{equation}
% where $\mathbf{x}_i^{(t-1)}$ is agent $i$'s accumulated context up to round $t-1$, $o_j^{(t-1)}$ denotes the full hybrid output of agent $j$ at the previous round, and $\oplus$ denotes concatenation over agents.

% \paragraph{Multi-Agent Co-Learning}
Based on the constructed inputs and outputs, we calculate the loss $L_{\text{stage1}}$ for each agent, then aggregate losses across all $N$ agents:
\begin{equation}
  \mathcal{L}_{\text{stage2}} = \frac{1}{N}\sum_{i=1}^{N}
  \Bigl(\alpha\,\mathcal{L}_{\text{hybrid}}^{(i)} + \beta\,\mathcal{L}_{\text{text}}^{(i)} + \gamma\,\mathcal{L}_{\text{align}}^{(i)}\Bigr),
  \label{eq:stage2loss}
\end{equation}
where $\mathcal{L}_{\text{hybrid}}^{(i)}$, $\mathcal{L}_{\text{text}}^{(i)}$, and 
$\mathcal{L}_{\text{align}}^{(i)}$ are respectively defined in Equations~\ref{eq:student}--\ref{eq:kd}.
% evaluated on agent $i$'s output at each round.
Through this joint training, agents are exposed to peers' latent vectors as part of their input context and learn to interpret them while generating a proper hybrid response.

%% file: latex/tex/3_exp_setup.tex
\section{Experimental Setup}\label{sec:setup}

\subsection{Training Data Construction}\label{sec:train_data}
% Ideally, training HyLaT would require large-scale, diverse multi-agent dialogue data covering a wide range of topics and communication styles, so that both the latent and text
% components generalise broadly. However, high-quality multi-agent conversational data with such structure and of this kind remains scarce. We therefore construct a limited subset as a proxy. 

\paragraph{Stage~1} To construct data required in Section~\ref{sec:stage1}, we collect datasets with detailed explanations or reasoning chains as supervision for the latent component: CommonsenseQA~\cite{talmor2019commonsenseqa},
SocialIQA~\cite{sap2019social}, WorldTree~\cite{jansen2018worldtree}, PubMedQA~\cite{jin2019pubmedqa}, and StrategyQA~\cite{geva2021did}. We pair each question with a detailed reasoning process and a concise final answer. This data-driven approach allows channel specialization to emerge naturally from the structure of the training examples, without requiring explicit routing annotations.

% All selected datasets contain detailed reasoning processes paired with concise final answers. This data-driven approach allows channel specialization to emerge naturally from the structure of the training examples, without requiring explicit routing annotations.

\paragraph{Stage~2} We construct two types of multi-agent interaction data. (1) \emph{Refinement}: using Llama-3.2-1B-Instruct and GPT-5~\cite{gpt5} to conduct simulated multi-agent debate on Stage~1 datasets, producing multi-round hybrid-output trajectories that converge to a final answer, where the answers of the first round may be incorrect. (2) \emph{Decomposition}: Multi-hop questions sampled from HotpotQA~\cite{yang2018hotpotqa}, 2WikiMultiHopQA~\cite{ho2020constructing}, and GSM8K-Aug-NL~\cite{shen2025codi} are decomposed into parallel sub-questions assigned across agents, who collaborate to reach a final answer. Dataset statistics and construction details are provided in Appendix~\ref{app:train_data}.

\subsection{Implementation Details}

We use Llama-3.2-1B-Instruct~\cite{grattafiori2024llama} as the backbone for main experiments, while experiments on other backbones are provided in Appendix~\ref{app:exp}. Parameter-efficient fine-tuning is applied via LoRA~\cite{hu2022lora} with rank $r=128$ and $\alpha=32$, applied to all attention projection matrices. The number of latent vectors per turn is set to $k=6$. Both stages are trained on 2-turn dialogues, with Stage~2 involving $N{=}2$ agents. All training experiments are conducted on 8 NVIDIA H200 GPUs. For evaluation, each task is run independently on a single H200 GPU. More details can be found in Appendix~\ref{app:imp_details}.

\input{latex/tabs/main_tmp}

\subsection{Evaluation}
\paragraph{Tasks}
Following \citet{du2023improving}, we adopt multi-agent debate (MAD) as the primary task for evaluation. We report results on both in-domain datasets encountered during training, i.e., CommonsenseQA, StrategyQA,
SocialIQA, WorldTree, and PubMedQA, as well as several out-of-domain datasets, specifically MedQA~\cite{jin2021disease}, ARC-Easy and ARC-Challenge~\cite{clark2018think}. We follow ~\citet{tang2025augmenting}
 to use the first 300 questions of each dataset to build MAD tasks. We conduct evaluations on these tasks with 3 agents communicating over 2 turns.
 
\paragraph{Metrics}
We assess both task performance and communication efficiency.
For \textbf{performance}, we report the average per-agent accuracy (Avg.) across all agents and the majority-vote accuracy (Maj.) aggregated over all agents. For \textbf{efficiency}, we report the average number of tokens generated per question across all
communication rounds (\# token), and the average running time per question.

\subsection{Compared Methods}
We compare HyLaT against various methods designed to optimize communication between agents. The ``\textbf{Single-Agent}'' baseline serves as a lower bound with no inter-agent communication. 
%For \textbf{multi-agent} baselines, we compare against methods that optimise the communication
%protocol in different ways:
Multi-agent baselines are further categorized into: (1) text-based methods, including:
\begin{itemize}[leftmargin=*, topsep=1pt]
\setlength{\itemsep}{0pt}
\setlength{\parsep}{0pt}
    \item \textbf{NL} uses unoptimised \underline{N}atural-\underline{L}anguage messages as the communication channel;
    \item \textbf{AutoForm}~\cite{chen2024beyond} directly prompts agents to communicate in structured language, such as JSON and code;
    \item \textbf{EcoLANG}~\cite{mou2025ecolang} evolves efficient communication rules via a selection procedure and prepends them to agent instructions;
    \item \textbf{SDE}~\cite{tang2025augmenting}, where agents transmit textual tokens alongside the model's internal
state-difference trajectory recorded in generation;
\end{itemize}
and (2) latent-space methods, including:
\begin{itemize}[leftmargin=*, topsep=1pt]
\setlength{\itemsep}{0pt}
\setlength{\parsep}{0pt}
    \item \textbf{Cipher}~\cite{pham2023let} replaces textual messages with semantic embedding vectors, computed as the belief-weighted average over the output vocabulary distribution.
    \item \textbf{LatentMAS-V\footnote{Since the original LatentMAS does not distinguish between dialogue roles and does not support multi-round multi-agent communication, we adapt its implementation to the multi-round
setting. This modification may not reflect its intended use and could underestimate its performance.}}~\cite{zou2025latent} 
% has each agent autoregressively generate hidden-state vectors
% from the final transformer layer and pass its full key--value cache as a latent working memory to the next agent.
has each agent autoregressively generate hidden-state vectors
from the final layer and pass its full key--value cache as a latent memory to the next agent.
\end{itemize}
Since the aforementioned methods do not involve additional training, we introduce additional baselines, \textbf{TextFullT} and \textbf{LatentFullT}, to ensure a fair comparison. \textbf{TextFullT} is trained on the same data as HyLaT but restricts all communication, including elaborations and conclusions to a single textual channel. \textbf{LatentFullT} is trained on the same data as HyLaT, but intermediate rounds communicate purely through latent vectors, while only the final round retains a latent vector and a text output for supervision and evaluation.

%% file: latex/tabs/main_tmp.tex
\begin{table*}[!t]
\centering
\setlength{\abovecaptionskip}{0.2cm}
\setlength{\belowcaptionskip}{-0.2cm}
\resizebox{\textwidth}{!}{
\begin{tabular}{@{}l|cccccccccc|cccccc|cc@{}}
\toprule
\multicolumn{1}{c|}{\multirow{3}{*}{\textbf{Method}}} & \multicolumn{10}{c|}{\textbf{In-Domain}} & \multicolumn{6}{c}{\textbf{Out-of-Domain}} &  \multicolumn{2}{|c}{\multirow{2}{*}{\textbf{\begin{tabular}[c]{@{}c@{}} Inference\\ Efficiency\end{tabular}}}} \\ \cmidrule{2-17}
\multicolumn{1}{c|}{} & \multicolumn{2}{c}{\textbf{Commonsense}} & \multicolumn{2}{c}{\textbf{StrategyQA}} & \multicolumn{2}{c}{\textbf{SocialIQA}} & \multicolumn{2}{c}{\textbf{WorldTree}} & \multicolumn{2}{c|}{\textbf{PubMedQA}} & \multicolumn{2}{c}{\textbf{MedQA}} & \multicolumn{2}{c}{\textbf{ARC-E}} & \multicolumn{2}{c|}{\textbf{ARC-C}} & \multicolumn{1}{c}{} \\ \cline{18-19} % \cmidrule(l){18-19} 
\multicolumn{1}{c|}{}                                 & Avg.            & Maj.           & Avg.            & Maj.          & Avg.           & Maj.          & Avg.           & Maj.          & Avg.           & Maj.         & Avg.         & Maj.        & Avg.         & Maj.        & Avg.          & Maj.        & \# token         & time        \\ \midrule
\multicolumn{10}{l}{\textit{\textbf{Training-Free Single-Agent Baseline}}} \\
Single-Agent                                                & 43.67          & 43.67          & 52.00          & 52.00         & 56.00         & 56.00         & 59.33         & 59.33         & 55.67         & 55.67        & 38.33       & 38.33       & 30.67       & 30.67       & 34.00        & 34.00       &      113.59    & \underline{1.25}                 \\ \midrule
\multicolumn{10}{l}{\textit{\textbf{Training-Free Text-Based Communication Methods}}} \\
NL                                                    & 49.56          & 53.33          & 54.44          & 54.67         & 58.22         & 61.00         & \underline{69.33}         & \underline{73.00}         & 54.56         & 59.00        & 42.67       & 42.67       & 35.33       & 37.67       & 31.78        & 32.00       &    1247.50        & 13.15                \\
AutoForm                                              & 47.78          & 50.33          & 54.56          & 55.00         & 59.89         & 63.00         & 67.11         & 68.67         & 56.11         & 57.00        & \underline{44.11}       & \textbf{46.00}       & 35.89       & 35.33       & 33.56        & 34.00       &     1717.60     & 18.18                \\
EcoLANG                                               & 48.00          & 50.67          & 49.44          & 50.33         & 55.78         & 61.33         & 68.11         & 71.33         & 54.67         & 57.33        & 41.89       & 42.67       & 37.33       & 36.67       & \textbf{34.89}        & \textbf{36.00}       &    960.26      & 10.04               \\
SDE                                                   & 47.22          & 49.00          & 52.67          & 52.67         & 58.11         & 60.67         & 64.56         & 68.00         & 49.67         & 53.00        & 40.78       & 41.67       & 34.78       & 36.67       & 32.89        & 32.33       &       1316.20   & 14.48                \\ \midrule
\multicolumn{10}{l}{\textit{\textbf{Training-Free Latent-Space Communication Methods}}} \\
Cipher                                                & 50.22          & 52.00          & 52.78          & 54.67         & 57.89         & 60.33         & 69.22         & 71.33         & 55.44         & 57.67        & 42.56       & 43.33       & 35.67       & 36.00       & 34.56        & 35.33       &  1137.89        & 11.58              \\
LatentMAS-V                                           & 26.11          & 26.00          & 49.67          & 50.00         & 42.78         & 43.00         & 45.00         & 46.33         & 54.22         & 55.00        & 32.67       & 33.00       & 27.67       & 27.89       & 27.00        & 28.33       &  324.43         & 3.61                 \\ \midrule
\multicolumn{10}{l}{\textit{\textbf{Training-Based Communication Methods}}} \\
TextFullT                                              & \textbf{64.44}          & \textbf{66.33}          & \underline{64.11}          & \underline{65.00}         & 76.66         & 77.63        & \textbf{70.56}         & \textbf{73.33}         & \textbf{69.67}         & \textbf{71.00}        & \textbf{44.55}       & \underline{44.67}       & \textbf{42.33}       & \textbf{42.67}       & \underline{34.89}        & \underline{34.67}       &     505.03     & 5.47                 \\
LatentFullT    &    56.00       &  56.33     &  63.11       &  63.33     &   \underline{80.78}      &   \underline{80.67}      &    68.56       &    68.33       &    66.11       &  66.67        &   42.11      &   42.33      &  38.89       &   38.67      &    33.78      &   34.00      &      \textbf{57.00}      &    \textbf{0.70}              \\
HyLaT                                                  & \underline{63.78}          & \underline{64.33}          & \textbf{64.67}          & \textbf{65.33}         & \textbf{82.56}         & \textbf{83.00}         & 69.22         & 69.00         & \underline{67.44}         & \underline{67.33}        & 42.11       & 42.33       & \underline{39.00}       & \underline{39.33}       & 33.89        & 34.00       &   \underline{72.01}         & 1.47                \\ \bottomrule
\end{tabular}
}
\caption{Experimental results of different methods. We report average accuracy (Avg.) and majority-vote 
accuracy (Maj.) across agents, along with communication 
efficiency measured by average token count (\# token) 
and wall-clock time (time, in seconds). 
\textbf{Bold} denotes the best result and \underline{underline} 
denotes the second best.}
\label{tab:exp_main}
\end{table*}

%% file: latex/tex/4_exp.tex
\section{Experimental Results}

\input{latex/tabs/ablation}

\subsection{Main Results}
Table~\ref{tab:exp_main} reports the overall performance of various methods, with several key findings highlighted:
% Based on these results, several insights can be gleaned:
%Overall, HyLaT achieves competitive 
% task performance across most tasks while dramatically reducing communication overhead.
\paragraph{I. Training-free methods fail to balance efficiency and performance.} Among training-free approaches: (1) The single-agent baseline achieves the highest efficiency but suffers from poor performance. (2) Text-based multi-agent methods yield strong results, yet offer limited room for efficiency optimization--even the most efficient one, EcoLANG, requires $8\times$ the duration of the single-agent baseline. (3) Latent-space methods are generally more efficient, particularly LatentMAS-V, but at the cost of a performance drop. These observations suggest that in a zero-shot setting, the fixed backbone significantly constrains agent behavior, highlighting the \textbf{necessity of additional training to co-optimize efficiency and performance}.

\paragraph{II. HyLaT strikes a balance between efficiency and performance.}
% This is demonstrated by: 
(1) \textbf{Unparalleled efficiency.} HyLaT operates at execution times close to the single-agent baseline while using 36.6\% fewer tokens. Specifically, HyLaT improves token and time efficiency by 13.3$\times$/6.8$\times$ and 4.5$\times$/2.5$\times$ over the most efficient text-based and latent-space baselines, respectively. (2) \textbf{Task performance.}  Unlike LatentMAS-V, HyLaT preserves strong task performance, achieving top-2 or comparative performance on both in-domain and out-of-domain datasets. Notably, on SocialIQA, HyLaT clearly outperforms all baselines by utilizing the latent channel to supplement information that text alone cannot fully convey. (3) \textbf{Advantage over TextFullT and LatentFullT.} Although HyLaT slightly trails TextFullT on tasks requiring exhaustive structured reasoning, such as WorldTree and ARC-E, its overall performance is highly comparable, while incurring substantially lower time and token costs. Meanwhile, LatentFullT only slightly improves computational efficiency, but its performance is inferior to that of HyLaT, suggesting that the intermediate text channel serves as a semantic anchor that maintains contextual coherence across rounds. 
% However, in terms of efficiency, the Explicit baseline incurs significantly higher time and token costs. 
In summary, HyLaT effectively enhances multi-agent communication efficiency, \textbf{maintaining task capabilities of other multi-agent frameworks while operating at single-agent costs}.

\subsection{Ablation Study}

% Table~\ref{tab:exp_ablation} presents ablation results averaged over all the tasks. We analyze the contribution of each component of HyLaT below.
Furthermore, we compare several variants of HyLaT in Table~\ref{tab:exp_ablation} to validate the effectiveness of the proposed dual-channel communication framework.

\paragraph{Effect of the Hybrid Channel Design}
We first ablate the channel composition by comparing two degenerate variants: (1) \textit{w/ pure text} uses our trained model, rather than TextFullT, to generate natural language throughout all rounds, resulting in significantly higher token usage and inference time, while yielding only marginal accuracy gains. This confirms that \textbf{the latent channel is the primary driver of HyLaT's efficiency advantage}. (2) \textit{w/ pure latent} restricts intermediate rounds to latent-only outputs, reserving hybrid generation for the final round only. Despite achieving the lowest token count, this variant suffers a notable drop in accuracy and a higher format error rate, suggesting that 
\textbf{textual signals are crucial for agents to maintain coherent comprehension across multiple turns}.
% the absence of natural-language signals in intermediate rounds 
% impairs agents' ability to maintain coherent comprehension across turns. 
% Together, these results validate the 
% design choice of maintaining both channels throughout all rounds of communication.
In summary, the advantages of the two channels are clearly complementary. Thus, the dual-channel design of HyLaT represents the optimal choice for balancing efficiency and performance.

\paragraph{Effect of Two-stage Training}
The variant \textit{w/o Stage~1} skips the single-agent warm-up and directly trains multi-agent interaction from scratch. Without first learning to generate stable hybrid outputs at the individual level, the latent representations produced early in training are noisy and unstable, and these errors propagate across rounds and agents, making optimization substantially more difficult and leading to negative results.
The variant \textit{w/o Stage~2} applies the Stage~1 checkpoint directly to multi-agent communication without co-learning. Since the model has never been exposed to peers' hybrid outputs during training, it frequently fails to process latent vectors from other agents at test time, resulting in a high format error rate (22.27\%) and a substantial drop in task performance. This confirms that Stage~2 co-learning is essential for enabling robust multi-round hybrid communication.

\subsection{Compatibility and Robustness of the Hybrid Protocol}

\begin{figure}[!t]
    \setlength{\abovecaptionskip}{-0.1cm}
    \setlength{\belowcaptionskip}{-0.4cm}
    \centering
    \includegraphics[width=\linewidth]{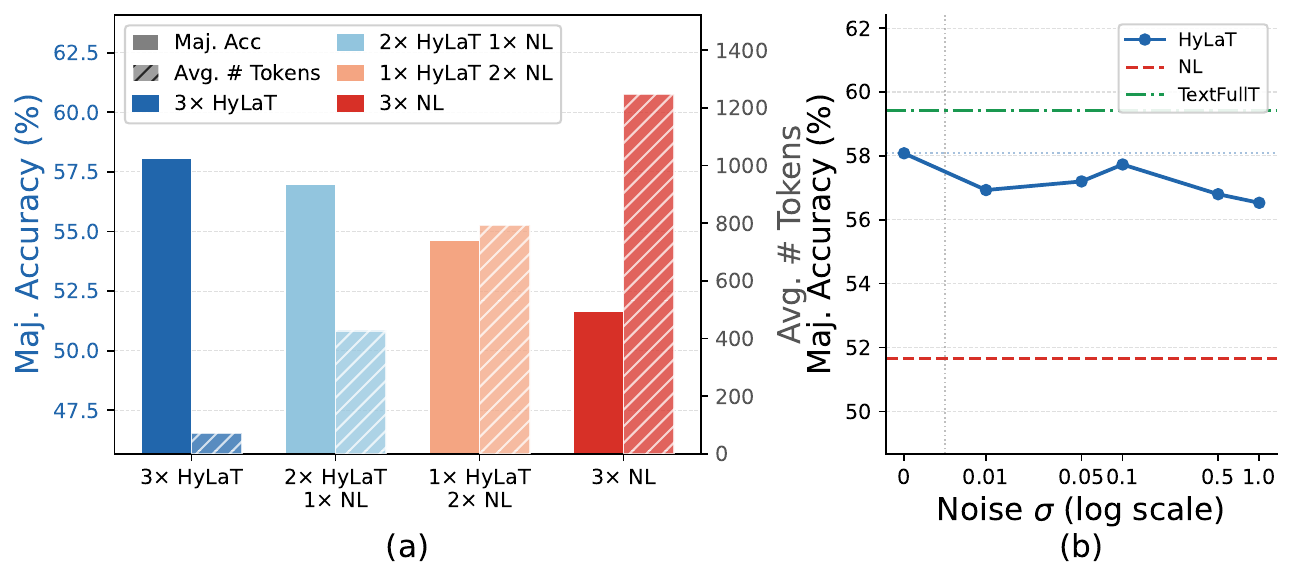}
    \caption{(a) Task performance and communication cost under heterogeneous agent settings; (b) Majority voting accuracy of HyLaT under increasing Gaussian noise applied to latent vectors at inference time. }
    \label{fig:exp_robust}
\end{figure}

\paragraph{Communication Protocol Compatibility}
A practical advantage of HyLaT's hybrid design is that the preserved text channel allows it to interoperate with agents using other communication protocols. To verify this, we construct heterogeneous multi-agent groups mixing HyLaT- and NL-based agents in varying proportions and evaluate task performance and token usage. As shown in Figure~\ref{fig:exp_robust}(a), increasing the proportion of HyLaT agents consistently reduces token consumption while improving task performance, indicating that HyLaT agents can effectively communicate with NL agents through the shared text channel and guide the group toward more efficient and accurate outcomes. This compatibility is unavailable to pure latent communication methods, whose opaque signals cannot be interpreted by agents operating in natural language.

\paragraph{Robustness to Latent Perturbation}

To assess the stability, we inject Gaussian noise of varying magnitude into each latent vector at inference time. As shown in Figure~\ref{fig:exp_robust}(b), HyLaT maintains stable performance for $\sigma \leq 0.5$.  This robustness stems from the hybrid channel design: while the latent channel carries intermediate reasoning, critical information such as the final answer is expressed through the text channel, where discrete tokens are more resilient to noise than continuous vectors, safeguarding the output from total degradation.
% whose discrete token generation is inherently more noise-tolerant than continuous vector representations. Perturbations to the latent channel therefore degrade reasoning quality only gradually, without catastrophically corrupting the agent's output.

\paragraph{Heterogeneous Agent Communication}
HyLaT can be naturally extended to heterogeneous multi-agent settings with minimal architectural modification. During Stage 2 training, a lightweight MLP adapter is introduced for each agent to project received latent representations into its own semantic space prior to fusion, thereby bridging the representational gap between agents from different model families. To validate this, we construct a heterogeneous variant of HyLaT pairing Llama-3.2-1B-Instruct and Qwen2-1.5B-Instruct as communicating agents. We evaluate this variant on the MAD task under a 2-agent, 2-round setting. As shown in Figure~\ref{fig:exp_hetero_scale}(a), the results demonstrate that HyLaT successfully enables effective communication between agents from different model families, confirming the generality of our hybrid latent-text communication framework.

\subsection{Scalability and Generalization Analysis}
% We evaluate HyLaT's generalization along two axes: scaling to larger agent populations and more interaction rounds, and transferring to a qualitatively different task domain.
This section demonstrates that HyLaT can scale to more complex agent interactions and generalizes to other task scenarios, such as social simulation.

\paragraph{Scalability of Agent Count and Interaction Rounds}
\begin{figure}[!t]
    \setlength{\abovecaptionskip}{-0.1cm}
    \setlength{\belowcaptionskip}{-0.4cm}
    \centering
    \includegraphics[width=\linewidth]{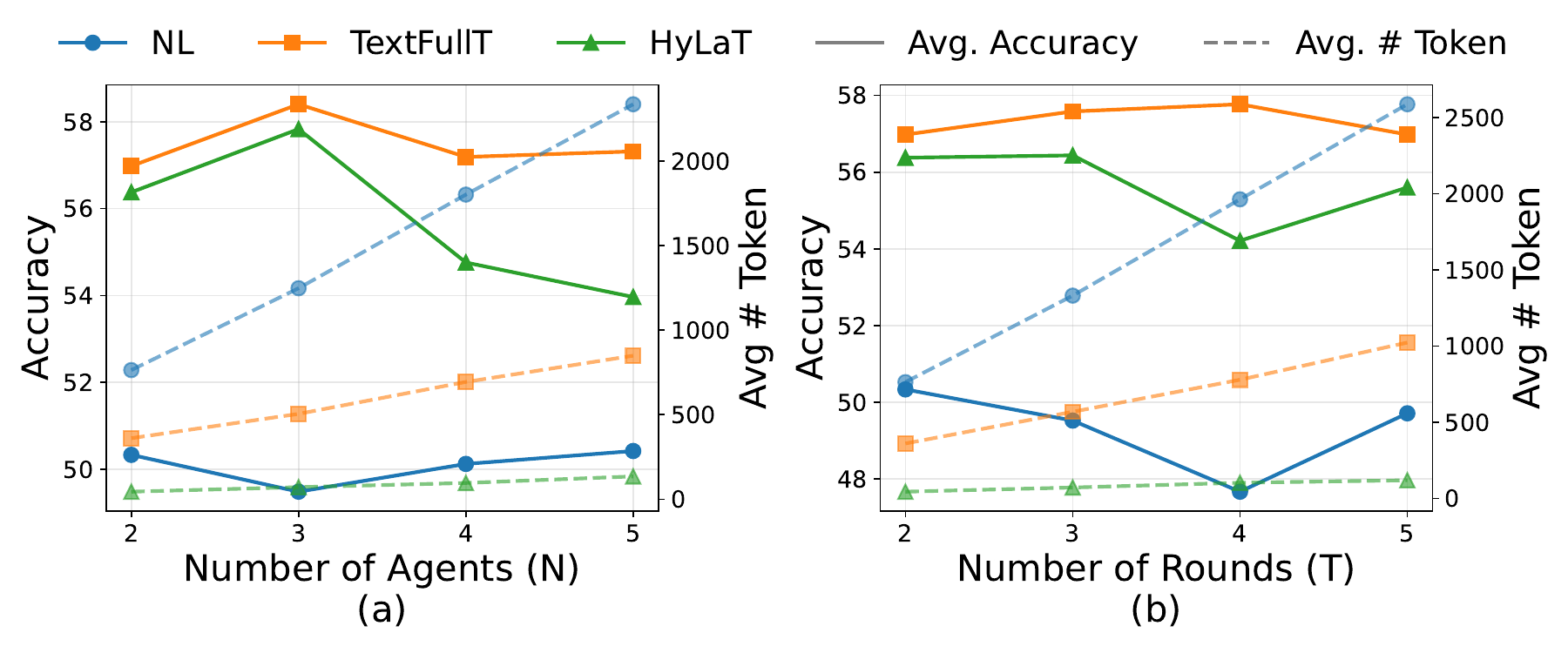}
    \caption{We vary the number of agents (N) while fixing the number of communication rounds to T=2 (left), and vary the number of rounds (T) while fixing N=2 (right).}
    \label{fig:exp_scale}
\end{figure}

As shown in Figure~\ref{fig:exp_scale}, HyLaT's token usage remains relatively stable as both $N$ and $T$ increase, while NL incurs rapidly growing overhead and TextFullT shows moderate growth. 
% On accuracy, all methods decline slightly with more agents or rounds, which we attribute to the limited capacity of the 1B backbone and the reduced output diversity after training rather than a failure of multi-agent communication. 
On accuracy, increasing $N$ or $T$ does not consistently lead to performance gains across all three methods, suggesting that effective coordination in more complex interactions generally remains challenging for 1B-scale models. For HyLaT, the degradation is also partially attributed to the gap between training and evaluation configuration: as the model is optimized for a small-scale setup where $N=2$, $T=2$, yet evaluated under larger-scale conditions. To verify this hypothesis, we constructed 3,400 training samples with 4-agent interactions and retrained Stage 2 of HyLaT. Figure~\ref{fig:exp_hetero_scale}(b) shows that the retrained model (HyLaT-scaled) notably alleviates the performance degradation. All these results demonstrate that \textbf{the efficiency advantage of HyLaT becomes increasingly pronounced as interactions scale, and that HyLaT can generalize to larger multi-agent settings, which can be further enhanced with correspondingly scaled training data}.

% Nevertheless, the magnitude of degradation remains moderate, indicating that HyLaT generalizes beyond its training configuration, with its \textbf{efficiency advantage becoming more pronounced as interactions scale}.

\paragraph{Generalization to Social Simulation}
\input{latex/tabs/social}
To further explore the applicability of HyLaT to multi-agent social simulation, we conduct the \textit{repeated trust game}
from~\citet{xie2024can}, in which a trustor and a trustee interact over 7 rounds: the trustor decides how much to send each round, and the trustee decides how much to return. As shown in Table~\ref{tab:exp_soc}, both HyLaT and TextFullT achieve substantially higher returned amounts than other methods. We attribute this to training on social reasoning data, which likely improves agents' ability to understand and respond to others' intentions. HyLaT additionally achieves this with far lower communication cost, suggesting that hybrid communication can preserve the social reasoning capability acquired while substantially reducing overhead.

% \begin{figure}[!t]
%     \setlength{\abovecaptionskip}{0cm}
%     \setlength{\belowcaptionskip}{-0.2cm}
%     \centering
%     \includegraphics[width=\linewidth]{latex/figs/attention.pdf}
%     % \caption{Example of top attended words at different generation positions of an agent when communicating with the other two agents. Tokens in black and \textcolor{softgray}{gray} represent words contained in questions and templates, respectively. \textcolor{softgreen}{Green} tokens and \textcolor{softblue}{blue} tokens represent the latent and text generated by the agent itself, and \textcolor{softorange}{orange} tokens and \textcolor{softred}{red} tokens are the latent and text generated by other agents.}
%     \caption{Top attended words at various generation steps during multi-agent communication. Tokens are color-coded by source: black (instructions and questions), \textcolor{softgray}{gray} (templates), \textcolor{softgreen}{green}/\textcolor{softblue}{blue} (agent’s own latent/text), and \textcolor{softorange}{orange}/\textcolor{softred}{red} (other agents’ latent/text).}
%     \label{fig:exp_attn}
% \end{figure}

\input{latex/tabs/attn}

\subsection{Interpretability Analysis}
\paragraph{Cross-Agent Attention Analysis}

To understand how agents utilize information from both channels during communication, we analyze the learned attention in Table~\ref{tab:attn}. In Round 1, latent generation primarily attends to keywords from the question and task instructions, while text generation shifts attention toward the agent's own latent tokens. In Round 2, a more striking pattern emerges: latent generation is dominated by the other agent's latent tokens from the previous round, providing suggestive evidence that cross-agent information exchange occurs through these continuous representations. When generating the critical textual output, the agent attends to a mixture of the other agent's textual response and its own latent tokens, indicating that the final decision integrates both explicitly communicated content and implicitly encoded details. 
Nevertheless, attention patterns alone do not represent causal explanations, and we will explore more rigorous interpretability analysis in future work.

% These results reveal a consistent two-stage flow within each round: the latent phase absorbs cross-agent signals, while the text phase synthesizes them into a response.

\paragraph{Probing the Latent Tokens}
To examine whether latent tokens encode meaningful information, we visualize the final-step latent embeddings across five datasets using PCA. As shown in Figure~\ref{fig:exp_pca}, the latent representations exhibit clear domain-level clustering despite the model never receiving explicit domain labels, demonstrating that the latent channel organizes information by semantic content rather than surface form. The first two principal components account for 70.5\% of the total variance, indicating that the latent space organizes around a small number of dominant semantic directions rather than distributing information uniformly. These observations suggest that latent tokens are not merely auxiliary padding: they actively encode task-specific semantics in a geometrically organized space, providing the downstream text decoder with structured, domain-relevant context.
% suggest that the latent tokens do capture domain-relevant structure, encoding task-specific semantics in a geometrically organized latent space. 
% Appendix~\ref{appendix:decoding} further explores whether the latent information can be decoded into texts through an external model.
% However, directly decoding the latent tokens into human-readable text remains challenging: we further apply the Sim-CoT explanation decoder~\cite{wei2025sim}, but it fails to produce meaningful reconstructions in our setting, which we attribute to the substantially richer information content that multi-agent debate requires each latent token to encode compared to the concise mathematical expressions in the original Sim-CoT setup.

%% file: latex/tabs/ablation.tex
\begin{table}[!t]
\centering
\setlength{\abovecaptionskip}{0.2cm}
\setlength{\belowcaptionskip}{-0.2cm}
\resizebox{0.48\textwidth}{!}{
\begin{tabular}{@{}l|ccccc@{}}
\toprule
\multicolumn{1}{c|}{\textbf{Method}} & \textbf{Avg.} & \textbf{Maj.} & \textbf{\# token} & \textbf{time} & \textbf{format err.(\%)} \\ \midrule
TextFullT & 58.40 & 59.41 & 505.03 & 5.47 & 0.00 \\
LatentFullT & 56.17 & 56.29 & 57.00 & 0.70 & 0.00 \\
Ours & 57.83 & 58.08 & 72.01 & 1.47 & 0.00 \\ \midrule
\multicolumn{6}{l}{\textbf{\textit{Variants based on Single-Channel Communication}}} \\
\textit{w/ pure text} & 56.93 & 57.00 & 639.18 & 13.18 & 3.28 \\
\textit{w/ pure latent} & 50.28 & 50.33 & 56.09 & 1.16 & 6.12 \\ \midrule
\multicolumn{6}{l}{\textbf{\textit{Variants with Single-Stage Training}}} \\
\textit{w/o Stage 1} & 38.43 & 38.58 & 102.59 & 1.38 & 9.13 \\
\textit{w/o Stage 2} & 43.64 & 43.96 & 136.10  & 2.87 & 22.27 \\ \bottomrule
\end{tabular}}
\caption{Ablation study of the HyLaT framework. Reported results are averaged over all datasets.}
\label{tab:exp_ablation}
\end{table}

%% file: latex/tabs/social.tex
\begin{table}[!t]
\centering
\setlength{\abovecaptionskip}{0.2cm}
\setlength{\belowcaptionskip}{-0.2cm}
\resizebox{0.45\textwidth}{!}{
\begin{tabular}{@{}l|cccc@{}}
\toprule
\multicolumn{1}{c|}{\multirow{2}{*}{\textbf{Method}}} & \multicolumn{4}{c}{\textbf{Trust}} \\ \cmidrule(l){2-5} 
\multicolumn{1}{c|}{} & Avg. Sent & Avg. Returned & \# token & time \\ \midrule
NL & 3.82 & 5.10 & 2134.13 & 22.36 \\
AutoForm & \textbf{4.56} & 7.75 & 4963.44 & 55.34 \\
EcoLANG & {\ul 4.05} & 7.13 & 776.81 & 8.43 \\
SDE & 3.66 & 6.60 & 1717.69 & 18.82 \\
TextFullT & 3.22 & {\ul 9.04} & 797.94 & 8.45 \\
HyLaT & 3.53 & \textbf{10.34} & 182.00 & 3.49 \\ \midrule
Human* & 7.48 & 12.24 & - & - \\ \bottomrule
\end{tabular}}
\caption{Results on \textit{repeated trust game} for human trust behavior simulation. \textit{Avg. Sent} and \textit{Avg. Returned} represent the average amount sent by the trustors and that returned by the trustees. Human*: estimated from figures in ~\cite{cochard2004trusting, xie2024can}}
\label{tab:exp_soc}
\end{table}

% % \useunder{\uline}{\ul}{}
% \begin{table*}[!t]
% \setlength{\belowcaptionskip}{-0.4cm}
% \resizebox{\textwidth}{!}{
% \begin{tabular}{@{}l|cccccccc@{}}
% \toprule
% \multicolumn{1}{c|}{\multirow{2}{*}{\textbf{Method}}} & \multicolumn{3}{c}{\textbf{Trust}} & \multicolumn{5}{c}{\textbf{Opinion}} \\ \cmidrule(l){2-9} 
% \multicolumn{1}{c|}{} & Avg. Sent & Avg. Returned & \# token & Bias(False) & Div.(False) & Bias(True) & Div.(True) & \# token \\ \midrule
% NL & 3.82 & 5.10 & 2134.13 & -0.51 & 0.72 & 0.60 & 0.67 & 2209.13 \\
% Autoform & \textbf{4.56} & 7.75 & 4963.44 & -0.16 & 0.55 & 0.04 & 0.68 & 3957.90 \\
% Ecolang & {\ul 4.05} & 7.13 & 776.81 & -0.18 & 0.63 & 0.57 & 0.57 & 768.90 \\
% SDE & 3.66 & 6.60 & 1717.69 & -0.36 & 0.89 & 0.42 & 0.70 & 1842.90 \\
% Explicit & 3.22 & {\ul 9.04} & 797.94 & -0.34 & 0.45 & 0.39 & 0.51 & 2077.30 \\
% HyLaT & 3.53 & \textbf{10.84} & 182.00 & -0.46 & 0.37 & 0.41 & 0.42 & 733.93 \\ \midrule
% Human & 7.48 & 12.24 & - & - & - & - & - & - \\ \bottomrule
% \end{tabular}}
% \caption{}
% \label{tab:soc}
% \end{table*}

%% file: latex/tabs/attn.tex
\begin{table}[!t]
\centering
\setlength{\abovecaptionskip}{0.2cm}
\setlength{\belowcaptionskip}{-0.2cm}
\small
\begin{tabular}{@{}p{0.17\linewidth}|p{0.80\linewidth}@{}}
\toprule
\multicolumn{2}{@{}p{0.98\linewidth}@{}}{%
\textbf{Example}
\newline
Answer the question based on the given context.\newline
Context: Jordan gave people hugs whether he knew them or not.\newline
Question: How would Others feel as a result?\newline
Options: A. like kissing him; B. uncomfortable; C. happy\newline
Briefly explain your answer, and write your final choice on ...} \\
\midrule
\textbf{Gen. Pos.} & \textbf{Top Attended Tokens} \\
 & {\footnotesize
\textcolor{black}{\rule{2mm}{2mm}}~instructions and questions\quad
\textcolor{softgray}{\rule{2mm}{2mm}}~template\quad
\textcolor{softgreen}{\rule{2mm}{2mm}}~an agent's own latent \quad
\textcolor{softblue}{\rule{2mm}{2mm}}~an agent's own text\quad
\textcolor{softorange}{\rule{2mm}{2mm}}~other agents' latent\quad
\textcolor{softred}{\rule{2mm}{2mm}}~other agents' text} \\
\midrule
Rd 1\newline 1st Latent &
\textcolor{softgray}{\texttt{<|begin\_of\_text|>}},
\textcolor{softgray}{\texttt{<bot>}},
parentheses, answer, context, example, uncomfortable, enclosed \\
Rd 1\newline Text Ans. &
\textcolor{softgray}{\texttt{<|begin\_of\_text|>}},
is,
\textcolor{softgreen}{\texttt{<latent\_0>}},
\textcolor{softgray}{\texttt{<eot>}},
answer, parentheses, B, uncomfortable,
\textcolor{softgray}{\texttt{<bot>}},
\textcolor{softgreen}{\texttt{<latent\_4>}},
\textcolor{softgreen}{\texttt{<latent\_5>}} \\
\midrule
Rd 2\newline 1st Latent &
\textcolor{softgray}{\texttt{<|begin\_of\_text|>}},
\textcolor{softgray}{\texttt{<bot>}},
answer,
\textcolor{softorange}{\texttt{<latent\_5>}},
\textcolor{softorange}{\texttt{<latent\_3>}},
\textcolor{softorange}{\texttt{<latent\_2>}},
\textcolor{softorange}{\texttt{<latent\_4>}},
\textcolor{softorange}{\texttt{<latent\_5>}},
\textcolor{softorange}{\texttt{<latent\_1>}},
parentheses,
\textcolor{softgreen}{\texttt{<latent\_4>}},
\textcolor{softgreen}{\texttt{<latent\_1>}},
\textcolor{softgreen}{\texttt{<latent\_2>}},
\textcolor{softgreen}{\texttt{<latent\_5>}} \\
Rd 2\newline Text Ans. &
\textcolor{softgray}{\texttt{<|begin\_of\_text|>}},
\textcolor{softgray}{\texttt{<eot>}},
answer, parentheses, A,
\textcolor{softblue}{B},
\textcolor{softblue}{B},
\textcolor{softgreen}{\texttt{<latent\_4>}},
\textcolor{softgreen}{\texttt{<latent\_5>}},
\textcolor{softgreen}{\texttt{<latent\_3>}},
\textcolor{softgreen}{\texttt{<latent\_2>}},
\textcolor{softred}{B},
\textcolor{softgreen}{\texttt{<latent\_1>}} \\
\bottomrule
\end{tabular}
\caption{Top attended words at various generation steps during multi-agent communication. Tokens are color-coded by source.}
\label{tab:attn}
\end{table}

%% file: latex/tex/5_related_work.tex
\section{Related Work}

\subsection{LLM-based Multi-agent Systems}

LLM-based multi-agent systems (MAS) have demonstrated strong performance across various scenarios~\cite{guo2024large,mou2024individual}. Representative applications include debate~\cite{du2023improving,tang2025augmenting}, software development~\cite{qian2024chatdev}, and social simulation~\cite{mou2024unveiling}. In these settings, agents communicate via natural language, structured messages, or shared memory to exchange information and coordinate actions. However, as the number of agents and interaction rounds increases, the rapidly growing communication overhead poses efficiency challenges~\cite{qian2024scaling,chen2025optima}.

\subsection{Multi-agent Communication}
To optimize the communication efficiency, early work on emergent communication~\cite{lazaridou2020emergent} showed that agents can develop their own compact protocols to solve cooperative tasks. With the rise of LLMs, recent studies have investigated how to optimize communication for both efficiency and performance. Some works~\cite{chen2024beyond,mou2025ecolang,chen2025optima} explore ways to compress or simplify message content while maintaining task effectiveness. Other works prune message-passing graph~\cite{zhang2024cut} or reorganize protocols~\cite{marro2024scalable} to reduce redundancy. Despite these advances, most approaches are limited to either structured text or learned symbolic languages, which constrains the density of information that can be exchanged.

\subsection{Latent Reasoning and Communication}

Recent work has explored latent representations as an alternative medium for reasoning in LLMs~\cite{hao2024training,shen2025codi,wei2025sim,dong2026interleaved}, improving efficiency compared to explicit chain-of-thought approaches. While this line of research mainly focuses on single-turn reasoning, it has recently been extended to multi-agent collaboration. These methods typically enable semantic transfer across models by sharing hidden states or projecting KV-cache representations~\cite{pham2023let,fu2025cache,zou2025latent,du2025enabling}. Despite these advances, existing approaches generally rely solely on latent communication, which reduces interpretability and controllability. Moreover, their thought-transfer mechanisms are often designed for sequential workflows and do not support multi-turn interactions, which are more common in general communication scenarios.

%% file: latex/tex/6_conclusion.tex
\section{Conclusion}
In this paper, we present HyLaT, a hybrid latent-text communication protocol that transfers elaborate cognitive signals through a latent channel and concise critical signals through a text channel. A two-stage training framework is proposed to enable agents to participate 
in effective multi-round hybrid communication. Experiments show that HyLaT dramatically improves communication efficiency while maintaining competitive task performance.

%% file: latex/tex/8_ack.tex
\section*{Acknowledgements}
The research is supported by National Key R\&D Program of China (Grant No. 2023YFF1204800) and the AI for Science Program, Shanghai Municipal Commission of Economy and Informatization (Grant No.2025-GZL-RGZN-BTBX-02028). The project’s computational resources are partially supported by CFFF platform of Fudan University.
YH was supported in part by the UK Engineering and Physical Sciences Research Council (EPSRC) through a Turing AI Fellowship (grant no. EP/V020579/1, EP/V020579/2).

In adherence to the ACL Publication Ethics Policy, we did not employ any AI assistant to generate the content of this paper. We only used AI assistants such as Claude to correct grammatical errors and enhance the writing quality.

%% file: latex/tex/7_limitations.tex
\section*{Limitations}
While HyLaT demonstrates promising results, several 
limitations remain. We discuss them below and 
outline directions for future work.
\begin{itemize}
    \item \textbf{Model scale.} Due to computational 
constraints, all experiments are conducted on relatively small models (1B and 3B parameters). Whether HyLaT's benefits extend to larger-scale models remains to be verified.
    \item \textbf{Deployment scale.} HyLaT's modified generation procedure is currently incompatible with inference frameworks such as vLLM, precluding evaluation in large-scale or long-horizon multi-agent scenarios, where communication efficiency becomes more important.
    \item \textbf{Text channel diversity.} The expressive diversity of the text channel is constrained by the scarcity of high-quality multi-agent communication data and human communication data.  Currently, our training set mainly includes QA-style debating data, and following~\citet{shen2025codi}, we append a fixed answer prompt to align generation, both of which limit the range of communicative behaviors the text channel can express. Exploring richer and more diverse communication scenarios is an important direction for future work.
    \item \textbf{Latent interpretability.} While HyLaT improves upon pure latent communication by preserving observable intermediate text outputs, the latent channel itself remains opaque. Decoding the semantic content of latent vectors is still highly challenging, and how to balance latent decodability with communication performance is a non-trivial open problem that we leave for future work.
\end{itemize}

%% file: latex/apps/app_data.tex
\section{Supplementary Implementation Details}

\subsection{Training Data Details}\label{app:train_data}

\paragraph{Datasets of Stage 1}
To support hybrid communication, Stage~1 training requires data that naturally exhibits the two-part structure of HyLaT's output: an elaborate cognitive signal encoding dense intermediate reasoning and a concise signal expressing the final response. We sample QA instances from existing datasets and pair them with detailed reasoning processes and short final answers. The reasoning processes are obtained from native annotations when available and from external~\cite{shen2025codi} or self-constructed annotations otherwise. In particular, SocialIQA is augmented with model-generated reasoning processes using GPT-5. Based on this, multiple QA pairs can be concatenated to form multi-turn dialogues. Dataset statistics are summarized in Table~\ref{tab:data_stage1}.
% Rather than constructing such data from scratch, we sample from training sets of existing datasets that already contain detailed reasoning processes alongside short final answers, treating the reasoning or explanation as the target for the latent channel and the concise answer as the supervision target for the text channel. Based on this, multiple QA pairs can be concatenated to form multi-turn dialogues. Dataset statistics are summarized in Table~\ref{tab:data_stage1}.

\paragraph{Datasets of Stage 2}
To support multi-round multi-agent communication, we construct training data through multi-agent debate simulation along two complementary axes. (1) \emph{Refinement}: In this setting, agents 
independently answer the same question in parallel and iteratively refine their responses through discussion. To ensure that inter-agent communication provides meaningful information gain, we deliberately avoid using a strong model for all rounds, as it tends to produce correct answers in the first round, leaving little room for correction through 
communication. Instead, we adopt a \emph{weak-strong pairing strategy}: the first round is generated by Llama-3.2-1B and the second round by GPT-5, and dialogues in which the second round answer remains incorrect are discarded. We randomly sample questions from the Stage~1 datasets and retain 3,960 valid 
dialogues; (2) \emph{Decomposition}: In this setting, agents collaboratively solve a complex question by first addressing complementary sub-problems in parallel, then synthesizing their findings to produce a final answer in the second round. We sample 2,000 structurally parallel questions from HotpotQA~\cite{yang2018hotpotqa}, 2WikiMultihopQA~\cite{ho2020constructing}, and GSM8K-AUG-NL~\cite{shen2025codi}, respectively. The prompts for data generation are provided in Figure~\ref{fig:prompt_data}.

\input{latex/tabs/data_stage1}

\input{latex/tabs/prompt_data}

\input{latex/tabs/llama_3b}

\input{latex/tabs/qwen}

\subsection{Evaluation Details}\label{app:eval_details}
We implement all the multi-agent communication experiments using the framework provided by~\citet{tang2025augmenting}. Following them, for datasets with more than 300 test examples, we evaluate on the first 300 samples. Evaluation tasks are constructed from the test splits of Stage~1 datasets as in-domain tasks, together with three additional out-of-domain datasets: MedQA~\cite{jin2021disease}, ARC-Easy and ARC-Challenge~\cite{clark2018think}. Unless otherwise specified, all main experiments are conducted with $N=3$ agents over $T=2$ interaction rounds. During inference, we use the following generation configuration: temperature $= 0.2$, top-$k$ $= 20$, top-$p$ $= 0.9$, with sampling enabled. The prompt templates for multi-agent debate are adapted from \citet{tang2025augmenting} (Figure~\ref{fig:prompt_eval}).
\input{latex/tabs/prompt_eval}

\subsection{Implementation 
Details}\label{app:imp_details}
\input{latex/tabs/hyper_param}
We follow previous work~\cite{hao2024training,shen2025codi} to set the number of latent vectors $k$ = 6, and we list the hyper-parameter settings used in Table~\ref{tab:hyper_param}. For Stage 2 training, we only supervise on the last turn, considering that the intermediate conclusions can be incorrect in the refinement data.

%% file: latex/tabs/data_stage1.tex
% Please add the following required packages to your document preamble:
% \usepackage{booktabs}
\begin{table}[!t]
\centering
\resizebox{0.48\textwidth}{!}{
\begin{tabular}{@{}l|l|l@{}}
\toprule
\multicolumn{1}{c|}{\textbf{Dataset}} & \multicolumn{1}{c|}{\textbf{Type}} & \multicolumn{1}{c}{\textbf{\# samples}} \\ \midrule
CommonsenseQA~\cite{talmor2019commonsenseqa} & commonsense reasoning & 3,600 \\
StrategyQA~\cite{geva2021did} & commonsense reasoning & 1,800 \\
SocialIQA~\cite{sap2019social} & social reasoning & 1,800 \\
WorldTree~\cite{jansen2018worldtree} & scientific question answering & 2,204 \\
PubMedQA~\cite{jin2019pubmedqa} & medical question answering & 500 \\ \bottomrule
\end{tabular}
}
\caption{Statistics of training datasets in Stage 1.}
\label{tab:data_stage1}
\end{table}

%% file: latex/tabs/prompt_data.tex
\begin{figure}[!t]
\centering
\begin{tcolorbox}[
    title={Round 1: Initial Prompt},
    fonttitle=\small
]
\small
Please answer the following question: \texttt{\{question\}}
\\
\\
First explain your reasoning, and provide your final answer in the form \texttt{\textbackslash boxed\{answer\}}, 
at the end of your response.
\end{tcolorbox}

\begin{tcolorbox}[
    title={Round $t > 1$: Multi-Agent Interaction Prompt},
    fonttitle=\small
]
\small
These are the solutions from other agents:
\\
One agent's response: \texttt{\`{}\`{}\`{}\{agent\_response\}\`{}\`{}\`{}}

[\textit{repeated for each other agent}]\\

Using the responses from other agents as additional information, can you provide your answer to the question? 
The original question is \texttt{\{question\}}. 
\\

Your final answer should be in the form 
\texttt{\textbackslash boxed\{answer\}}, at the end 
of your response.
\end{tcolorbox}
\caption{Prompts used for multi-agent interaction data construction. 
Round 1 elicits an initial response with explicit reasoning. 
In subsequent rounds, each agent receives the responses of all other agents and is asked to produce a revised answer.}
\label{fig:prompt_data}
\end{figure}

%% file: latex/tabs/llama_3b.tex
\begin{table*}[!t]
\centering
\setlength{\abovecaptionskip}{0.2cm}
\resizebox{\textwidth}{!}{
\begin{tabular}{@{}l|cccccccccc|cccccc|cc@{}}
\toprule
\multicolumn{1}{c|}{\multirow{3}{*}{\textbf{Method}}} & \multicolumn{10}{c|}{\textbf{In-Domain}} & \multicolumn{6}{c|}{\textbf{Out-of-Domain}} & \multicolumn{2}{c}{\multirow{2}{*}{\textbf{\begin{tabular}[c]{@{}c@{}}Inference\\ Efficiency\end{tabular}}}} \\ \cmidrule{2-17}
\multicolumn{1}{c|}{} & \multicolumn{2}{c}{\textbf{Commonsense}} & \multicolumn{2}{c}{\textbf{StrategyQA}} & \multicolumn{2}{c}{\textbf{SocialIQA}} & \multicolumn{2}{c}{\textbf{WorldTree}} & \multicolumn{2}{c|}{\textbf{PubMedQA}} & \multicolumn{2}{c}{\textbf{MedQA}} & \multicolumn{2}{c}{\textbf{ARC-E}} & \multicolumn{2}{c|}{\textbf{ARC-C}} & \multicolumn{1}{c}{} \\ \cline{18-19}
\multicolumn{1}{c|}{} & Avg. & Maj. & Avg. & Maj. & Avg. & Maj. & Avg. & Maj. & Avg. & Maj. & Avg. & Maj. & Avg. & Maj. & Avg. & Maj. & \# token & time \\ \midrule
\multicolumn{19}{l}{\textit{\textbf{Training-Free Single-Agent Baseline}}} \\
Single-Agent & 73.67 & 73.67 & 66.00 & 66.00 & 74.00 & 74.00 & \uline{89.00} & 89.00 & 71.00 & 71.00 & 62.00 & 62.00 & 42.67 & 42.67 & \uline{45.67} & 45.67 & 130.65 & 2.46 \\ \midrule
\multicolumn{19}{l}{\textit{\textbf{Training-Free Text-Based Communication Methods}}} \\
NL          & 72.33 & 76.00 & 66.56 & 67.67 & 74.89 & 76.33 & \textbf{89.78} & \textbf{90.33} & 69.00 & 69.00 & 60.44 & 62.00 & 44.00 & \textbf{44.67 }& 44.89 & 46.00 & 1094.83 & 19.13 \\
AutoForm    & 68.11 & 69.00 & 70.33 & 72.67 & 75.56 & 76.00 & 88.78 & \uline{90.00} & 69.56 & 69.33 & \textbf{65.00} & \textbf{68.00} & 42.56 & 43.00 & 44.22 & 44.33 & 1668.19 & 29.09 \\
EcoLANG     & 72.33 & 74.33 & 68.44 & 68.00 & 74.44 & 73.33 & 87.89 & 88.33 & 69.22 & 69.33 & \uline{61.11} & 61.33 & 43.00 & 43.67 & 42.78 & 42.00 & 603.64 & 10.43 \\
SDE         & 72.56 & 74.00 & 64.89 & 65.33 & 72.89 & 74.67 & 86.56 & 88.67 & 69.00 & 69.00 & 58.67 & \uline{62.67} & 42.56 & 43.67 & 43.67 & 44.67 & 913.33 & 16.35 \\ \midrule
\multicolumn{19}{l}{\textit{\textbf{Training-Free Latent-Space Communication Methods}}} \\
Cipher      & 71.67 & 73.67 & 66.89 & 69.33 & 74.89 & 75.00 & 86.56 & 87.00 & 67.89 & 69.33 & 59.56 & 60.33 & \textbf{44.11} & 44.33 & \textbf{46.11} & \textbf{48.00} & 1080.19 & 18.40 \\
LatentMAS-V & 70.78 & 72.17 & 59.67 & 61.67 & 73.00 & 74.33 & 84.56 & 84.67 & 62.44 & 64.00 & 60.11 & 61.67 & 41.78 & 41.33 & 42.89 & 44.33 & 570.63 & 10.58 \\ \midrule
\multicolumn{19}{l}{\textit{\textbf{Training-Based Communication Methods}}} \\
TextFullT    & 72.22 & 72.67 & 77.22 & 78.67 & 84.78 & 84.33 & 87.67 & 89.33 & \textbf{74.00} & \textbf{75.33} & 60.44 & 62.00 & 43.22 & 43.33 & 44.89 & \uline{46.00} & 748.74 & 12.79 \\
LatentFullT    & \uline{78.67} & \uline{78.33} & \uline{81.22} & \uline{81.00} & \uline{89.22} & \uline{89.67} & 84.67 & 84.67 & 73.56 & 73.33 & 59.44 & 59.00 & 43.67 & 43.67 & 43.33 & 43.33 & \textbf{57.00} & \textbf{1.21} \\
HyLaT       & \textbf{78.67} & \textbf{78.33} & \textbf{85.00} & \textbf{85.00} & \textbf{90.11} & \textbf{90.00} & 86.89 & 86.33 & \uline{73.89} & \uline{74.00} & 59.44 & 59.33 & \uline{44.00} & \uline{44.00} & 43.44 & 43.67 & \uline{72.00} & \uline{2.46} \\ \bottomrule
\end{tabular}
}
\caption{Experimental results of different methods on Llama-3.2-3B-Instruct. We report average accuracy (Avg.) and majority-vote accuracy (Maj.) across agents, along with communication efficiency measured by average token count (\# token) and wall-clock time (time, in seconds). \textbf{Bold} denotes the best result and \underline{underline} denotes the second best.}
\label{tab:exp_3b}
\end{table*}

%% file: latex/tabs/qwen.tex
\begin{table*}[!t]
\centering
\setlength{\abovecaptionskip}{0.2cm}
\resizebox{\textwidth}{!}{
\begin{tabular}{@{}l|cccccccccc|cccccc|cc@{}}
\toprule
\multicolumn{1}{c|}{\multirow{3}{*}{\textbf{Method}}} & \multicolumn{10}{c|}{\textbf{In-Domain}} & \multicolumn{6}{c|}{\textbf{Out-of-Domain}} & \multicolumn{2}{c}{\multirow{2}{*}{\textbf{\begin{tabular}[c]{@{}c@{}}Inference\\ Efficiency\end{tabular}}}} \\ \cmidrule{2-17}
\multicolumn{1}{c|}{} & \multicolumn{2}{c}{\textbf{Commonsense}} & \multicolumn{2}{c}{\textbf{StrategyQA}} & \multicolumn{2}{c}{\textbf{SocialIQA}} & \multicolumn{2}{c}{\textbf{WorldTree}} & \multicolumn{2}{c|}{\textbf{PubMedQA}} & \multicolumn{2}{c}{\textbf{MedQA}} & \multicolumn{2}{c}{\textbf{ARC-E}} & \multicolumn{2}{c|}{\textbf{ARC-C}} & \multicolumn{1}{c}{} \\ \cline{18-19}
\multicolumn{1}{c|}{} & Avg. & Maj. & Avg. & Maj. & Avg. & Maj. & Avg. & Maj. & Avg. & Maj. & Avg. & Maj. & Avg. & Maj. & Avg. & Maj. & \# token & time \\ \midrule
\multicolumn{19}{l}{\textit{\textbf{Training-Free Single-Agent Baseline}}} \\
Single-Agent & 58.00 & 58.00 & 50.67 & 50.67 & 68.67 & 68.67 & 72.00 & 72.00 & \textbf{61.33} & \textbf{61.33} & \textbf{39.00} & 39.00 & 38.00 & 38.00 & 35.00 & 35.00 & 87.11 & \uline{1.65} \\ \midrule
\multicolumn{19}{l}{\textit{\textbf{Training-Free Text-Based Communication Methods}}} \\
NL          & 59.33 & 60.67 & 50.33 & 50.00 & 67.22 & 69.00 & 74.33 & 77.00 & 58.11 & 58.00 & 37.78 & 38.33 & 37.67 & 38.33 & 38.00 & 39.67 & 552.40 & 10.38 \\
AutoForm    & 50.78 & 55.33 & 50.22 & 50.67 & 66.56 & 68.00 & 77.89 & \textbf{79.67} & 56.78 & 57.33 & 35.33 & 36.33 & 37.67 & 38.33 & 35.33 & 36.67 & 344.69 & 6.51 \\
EcoLANG     & 61.11 & 62.00 & 51.00 & 51.00 & 69.22 & 69.33 & \uline{78.56} & 78.67 & 59.00 & 58.67 & 36.22 & 36.67 & 36.67 & 37.33 & \textbf{39.67} & \uline{39.67} & 235.55 & 4.49 \\
SDE         & 58.67 & 61.33 & 49.67 & 49.33 & 67.56 & 70.33 & 75.44 & 78.33 & 59.44 & 59.67 & \uline{38.78} & \textbf{40.67} & 38.00 & 39.33 & 37.22 & 38.33 & 1514.63 & 10.29 \\ \midrule
\multicolumn{19}{l}{\textit{\textbf{Training-Free Latent-Space Communication Methods}}} \\
Cipher      & 60.11 & 60.00 & 51.67 & 51.67 & 69.78 & 70.00 & 78.33 & 78.67 & \uline{61.11} & 61.00 & 36.00 & 36.33 & 39.33 & 39.33 & 36.11 & 36.33 & 691.38 & 7.84 \\
LatentMAS-V & 4.56 & 4.00 & 0.33 & 0.00 & 13.78 & 13.00 & 10.33 & 10.00 & 21.00 & 20.33 & 13.89 & 12.33 & 4.33 & 3.67 & 4.33 & 3.00 & 587.99 & 11.86 \\ \midrule
\multicolumn{19}{l}{\textit{\textbf{Training-Based Communication Methods}}} \\
TextFullT   & \uline{62.56} & \uline{62.00} & 55.11 & 55.00 & 76.11 & 76.00 & 78.33 & 78.67 & 59.11 & 60.33 & 38.67 & \uline{39.00} & 38.56 & 40.00 & 37.67 & 39.33 & 545.30 & 10.06 \\
LatentFullT & 59.83 & 60.00 & \uline{55.17} & \uline{56.67} & \uline{76.17} & \uline{76.33} & 76.00 & 75.33 & 53.67 & 53.33 & 36.50 & 38.33 & \uline{41.00} & \uline{41.00} & 37.67 & 38.67 & \textbf{46.00} & \textbf{0.81} \\
HyLaT       & \textbf{62.78} & \textbf{63.00} & \textbf{55.89} & \textbf{56.67} & \textbf{76.44} & \textbf{76.33} & \textbf{78.56} & \uline{78.67} & 59.67 & \uline{61.00} & 38.67 & 38.67 & \textbf{41.89} & \textbf{41.67} & \uline{39.44} & \textbf{39.67} & \uline{66.00} & 2.43 \\ \bottomrule
\end{tabular}
}
\caption{Experimental results of different methods on Qwen2-1.5B-Instruct. We report average accuracy (Avg.) and majority-vote accuracy (Maj.) across agents, along with communication efficiency measured by average token count (\# token) and wall-clock time (time, in seconds). \textbf{Bold} denotes the best result and \underline{underline} denotes the second best.}
\label{tab:exp_qwen}
\end{table*}

%% file: latex/tabs/prompt_eval.tex
\begin{figure}[!t]
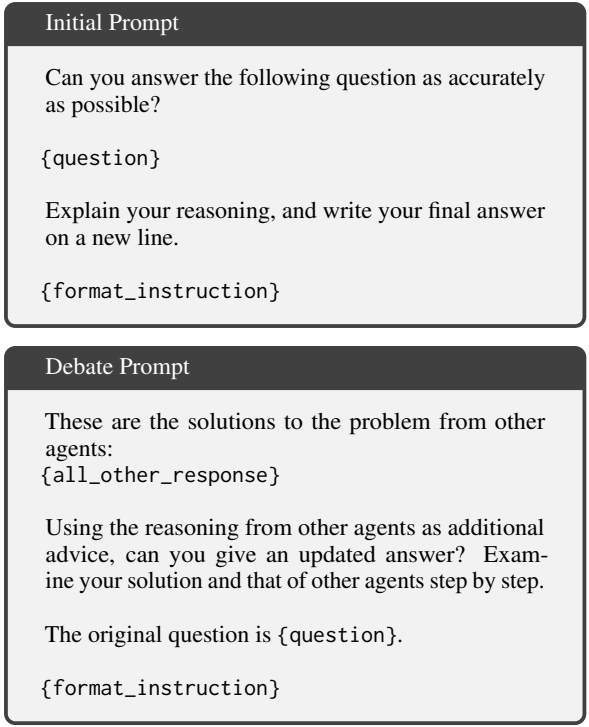

\centering
\begin{tcolorbox}[
    title={Initial Prompt},
    fonttitle=\small
]
\small
Can you answer the following question as accurately as possible?
\\

\texttt{\{question\}}\\

Explain your reasoning, and write your final answer 
on a new line.\\

\texttt{\{format\_instruction\}}
\end{tcolorbox}

\begin{tcolorbox}[
    title={Debate Prompt},
    fonttitle=\small
]
\small
These are the solutions to the problem from other 
agents:

\texttt{\{all\_other\_response\}}\\

Using the reasoning from other agents as additional 
advice, can you give an updated answer? Examine your solution and that of other agents step by step.\\

The original question is \texttt{\{question\}}.\\

\texttt{\{format\_instruction\}}
\end{tcolorbox}
\caption{Prompt templates used for multi-agent debate evaluation.}
\label{fig:prompt_eval}
\end{figure}

%% file: latex/tabs/hyper_param.tex
% Please add the following required packages to your document preamble:
% \usepackage{booktabs}
\begin{table}[!t]
\centering
\resizebox{0.48\textwidth}{!}{
\begin{tabular}{@{}lcc@{}}
\toprule
\multicolumn{1}{c}{\textbf{Configuration}} & \textbf{Stage 1} & \textbf{Stage 2} \\ \midrule
Model initialization & Llama-3.2-1B-Instruct & Stage 1 \\
LoRA & rank=128, alpha=32 & rank=128, alpha=32 \\
Global batch size & 128 & 128 \\
Peaking learning rate & 8.00e-04 & 3.00e-04 \\
Optimizer & AdamW & AdamW \\
LR scheduler & Cosine & Cosine \\
Training epochs & 3 & 3 \\
Warmup ratio & 0.03 & 0.03 \\
Precision & bfloat16 & bfloat16 \\
$\alpha$ & 1 & 1 \\
$\beta$ & 1 & 1 \\
$\gamma$ & 20 & 20 \\ \bottomrule
\end{tabular}
}
\caption{The detailed training hyper-parameters of HyLaT.}
\label{tab:hyper_param}
\end{table}

%% file: latex/apps/app_exp.tex
\section{Supplementary Experimental Results and Analysis}\label{app:exp}

\subsection{Heterogeneous Agent Communication}
\begin{figure}[!t]
    \setlength{\abovecaptionskip}{-0.1cm}
    \setlength{\belowcaptionskip}{-0.4cm}
    \centering
    \includegraphics[width=\linewidth]{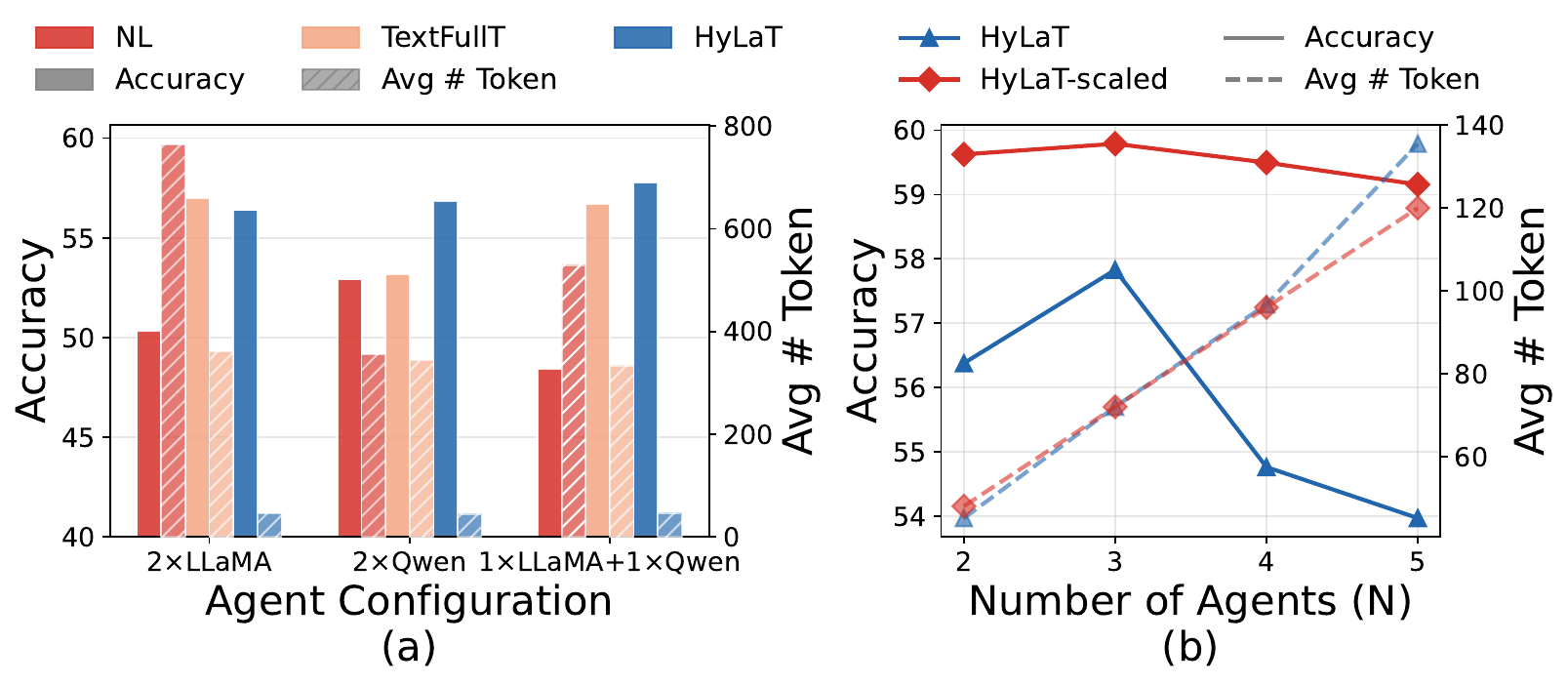}
    \caption{(a) Performance of different communication methods under homogeneous and heterogeneous agent configurations; (b) Performance of HyLaT trained on original vs. scaled data as the number of agents increases.}
    \label{fig:exp_hetero_scale}
\end{figure}

We implemented a heterogeneous variant of HyLaT pairing Llama-3.2-1B and Qwen2-1.5B as communicating agents, with a lightweight MLP adapter bridging their respective latent spaces. We evaluated this variant on the MAD task under a 2-agent and 2-round setting. Figure~\ref{fig:exp_hetero_scale}(a) shows that although the heterogeneous setting sometimes does not surpass the best homogeneous configuration (2 × stronger model), HyLaT consistently performs well across all agent configurations.

\subsection{Further Improve the Scalability}
In Figure~\ref{fig:exp_scale}(a), we can observe that performance begins to degrade when the number of agents exceeds 3. This degradation with more agents is attributed to two factors: (1) this phenomenon is a general characteristic of multi-agent systems at this model scale or for these tasks, rather than a limitation introduced by our method alone, as TextFullT exhibits the same degradation pattern when the number of agents exceeds 3. (2) the degradation in HyLaT is partly due to a gap between training and evaluation: Our current training setup uses 2-agent interaction, which does not fully cover the distribution of larger-scale agent configurations at evaluation time. To verify this hypothesis and mitigate the issue, we constructed 3,400 training samples with 4-agent interactions using the \textit{refinement} method in Sec~\ref{sec:train_data}, and retrained stage 2 of HyLaT. Results in Figure~\ref{fig:exp_hetero_scale}(b) show that such training data notably alleviates the performance degradation, confirming that the scalability limitation stems from insufficient training coverage rather than a fundamental constraint of our method. 

\subsection{Probing the Latent Tokens}

% \begin{figure}[!t]
%     \setlength{\abovecaptionskip}{0.2cm}
%     \centering
%     \includegraphics[width=\linewidth]{latex/figs/pca.pdf}
%     \caption{PCA analysis of final-step latent embeddings across five datasets.}
%     \label{fig:exp_pca}
% \end{figure}

% \begin{figure}[!t]
%     \setlength{\abovecaptionskip}{0.2cm}
%     \centering
%     \includegraphics[width=\linewidth]{latex/figs/pca_scree.pdf}
%     \caption{Explained variance ratio of the top-10 principal components of the latent embeddings.}
%     \label{fig:exp_pca_scree}
% \end{figure}

\begin{figure}[!t]
    \setlength{\abovecaptionskip}{0.2cm}
    \centering
    \includegraphics[width=\linewidth]{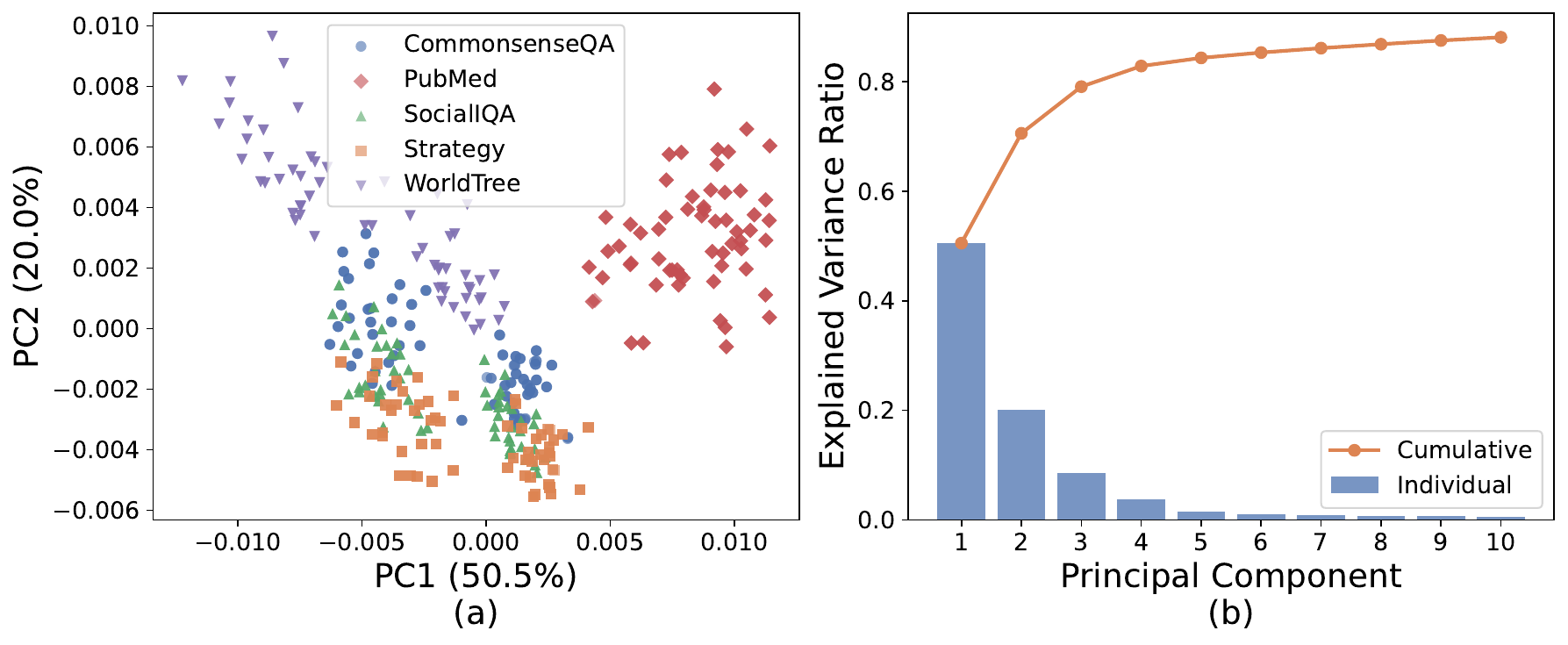}
    \caption{(a) PCA analysis of final-step latent embeddings across five datasets. (b) Explained variance ratio of the top-10 principal components of the latent embeddings.}
    \label{fig:exp_pca}
\end{figure}

To probe the semantic content encoded in HyLaT's latent channel, we collect 180 latent embeddings produced by agents at the final interaction round for each for the five datasets and apply PCA to project them into two dimensions. The results are visualized in Figure~\ref{fig:exp_pca}. As shown in the figure, embeddings from the same  dataset form relatively compact clusters that are well-separated from those of other datasets, suggesting that the latent vectors capture dataset-specific semantic structure rather than collapsing into an undifferentiated representation. This provides indirect evidence that the latent channel encodes meaningful content that reflects the nature of the task and communicative context, even though the vectors are never decoded into discrete tokens during inference.

% \subsubsection{Decoding Latent Tokens into Texts}
% \label{appendix:decoding}
% However, directly decoding the latent tokens into human-readable text remains challenging: we further apply the Sim-CoT explanation decoder~\cite{wei2025sim}, but it fails to produce meaningful reconstructions in our setting, which we attribute to the substantially richer information content that multi-agent debate requires each latent token to encode compared to the concise mathematical expressions in the original Sim-CoT setup.

\subsection{Different Model Scales}\label{app:exp_3b}
We extend the experiments on larger Llama backbones, i.e., Llama-3.2-3B-Instruct. Results in Table~\ref{tab:exp_3b} show that HyLaT scales effectively to this setting: task performance remains competitive across most benchmarks, and the efficiency advantage is fully preserved. The exception is WorldTree, where the 3B model already achieves strong performance under text-based communication, leaving limited room for further improvement.

\subsection{Different Model Families}\label{app:qwen}
We also conducted additional experiments using Qwen2-1.5B-Instruct as the backbone. Results in Table~\ref{tab:exp_qwen} show that HyLaT also demonstrates effectiveness on Qwen, matching or outperforming TextFullT across most tasks, while achieving dramatically higher communication efficiency. We also note that LatentMAS-V fails to produce meaningful results on most MAD tasks on Qwen, suggesting that adapting the original approach to multi-round multi-agent communication for different models requires non-trivial effort. Overall, the aforementioned results demonstrate that HyLaT possesses superior cross-family generalization.